\documentclass[pdflatex,sn-mathphys-ay]{sn-jnl}

\usepackage{pdflscape}
\usepackage{graphicx}
\usepackage{bm}
\newtheorem{definition}{Def}[section]
\newtheorem{theorem}{Theo}[section]

\usepackage[cmyk]{xcolor} 
\usepackage{lipsum}
\usepackage{lineno}

\usepackage{booktabs}

\usepackage{multirow}
\usepackage{array}

\usepackage{hyperref}
\usepackage{amsmath}

\begin{document}



\title{Predicting Subway Passenger Flows under Incident Situation with Causality } 


\author[1]{\fnm{Xiannan} \sur{Huang} 
}

\author[1]{\fnm{Shuhan} \sur{Qiu} 
}

\author[2]{\fnm{Quan} \sur{Yuan}
}
\author*[1,2]{\fnm{Chao} \sur{Yang} \email{tongjiyc@tongji.edu.cn}}

\affil[1]{\orgname{Key Laboratory of Road and Traffic Engineering, Ministry of Education at Tongji University}, \orgaddress{\street{4800 Cao’an Road}, \city{Shanghai}, \postcode{201804}, \state{Shanghai}, \country{China }}}

\affil[2]{ \orgname{Urban Mobility Institute, Tongji University}, \orgaddress{\street{1239 Siping Road}, \city{Shanghai}, \postcode{200092}, \state{Shanghai}, \country{China }}}


\abstract{In the context of rail transit operations, real-time passenger flow prediction is essential; however, most models primarily focus on normal conditions, with limited research addressing incident situations. There are several intrinsic challenges associated with prediction during incidents, such as a lack of interpretability and data scarcity. To address these challenges, we propose a two-stage method that separates predictions under normal conditions and the causal effects of incidents. First, a normal prediction model is trained using data from normal situations. Next, the synthetic control method is employed to identify the causal effects of incidents, combined with placebo tests to determine significant levels of these effects. The significant effects are then utilized to train a causal effect prediction model, which can forecast the impact of incidents based on features of the incidents and passenger flows. During the prediction phase, the results from both the normal situation model and the causal effect prediction model are integrated to generate final passenger flow predictions during incidents. Our approach is validated using real-world data, demonstrating improved accuracy. Furthermore, the two-stage methodology enhances interpretability. By analyzing the causal effect prediction model, we can identify key influencing factors related to the effects of incidents and gain insights into their underlying mechanisms. Our work can assist subway system managers in estimating passenger flow affected by incidents and enable them to take proactive measures. Additionally, it can deepen researchers' understanding of the impact of incidents on subway passenger flows.}



\maketitle
\section{Introduction}
The subway system is a critical component of the transportation network in metropolitan areas due to its environmental benefits, efficiency, and reliability. Understanding passenger flow patterns within these systems is important for enhancing operational efficiency and user satisfaction. While much research has focused on typical operating conditions, there is a significant gap in addressing unusual situations, such as incidents. Analyzing and predicting how incidents impact passenger flows is vital for subway operators to implement proactive strategies. For instance, some studies about emergency bus bridging and dispatching models during subway incidents use the affected passenger numbers as inputs for their models \cite{Zhang2023Resi,Tan2024EnhancingUM}. Consequently, we aim to fill this gap in our paper by proposing a method to \textit{interpret the influence of incident on OD (origination- destination) passenger flows and predict the passenger flows under incident situation}.

There are several papers \cite{Zou2024RealtimePO,NOURSALEHI2018277} that focus on predicting passenger flows in subway systems, treating incidents as input features. Since incidents are characteristics of the inputs, these models can predict passenger flows under both normal and incident conditions. They demonstrate improved predictive power and provide valuable insights. However, these models employ complex deep learning methods and end-to-end structures, which might obscure the mechanisms by which incidents affect passenger flows. Furthermore, it can be challenging for these models to learn patterns related to incidents in depth, as the number of incidents in subway systems is typically small. Additionally, the imbalance between the volume of data in normal and incident situations may lead the model to prioritize normal conditions, hindering its ability to adequately learn the rules governing incident situations. Finally, the criteria for determining which OD flows are influenced by incidents have not been thoroughly discussed. Some methods \cite{Zou2024RealtimePO}
assume that the influenced OD flows only occur during the incident period. But this assumption is too simple and contradicts reality, for example, even though an incident ends, the effects of it may still persist. And we will elaborate it in the experiment section. 

\begin{figure}
    \centering
    \includegraphics[width=\linewidth]{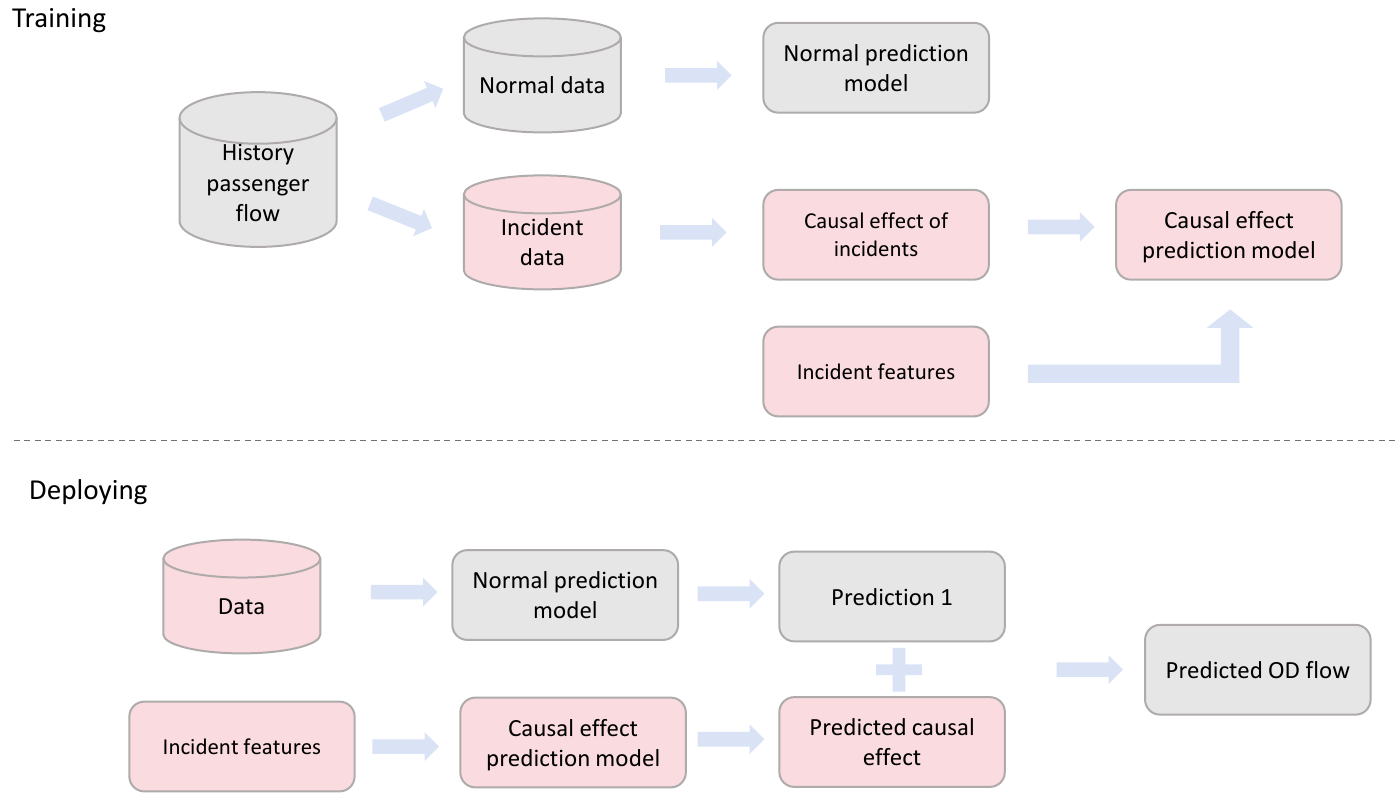}
    \caption{The workflow of our prediction method}
    \label{fig:wolkflow-all_1}
\end{figure}

To address these gaps, we propose a two-stage method for predicting passenger flow during incidents, as illustrated in Figure \ref{fig:wolkflow-all_1}. First, a model trained with passenger flow data under normal conditions can be utilized to predict the OD flows that would occur in the absence of an incident. Subsequently, another model can be employed to predict the causal effects of the incident on passenger flows. These causal effects can then be incorporated into the normal OD flow predictions to obtain the final predictions. Besides, these causal effects can also be fed into emergency bus bridging and dispatching models \cite{Zhang2023Resi,Tan2024EnhancingUM} to generate an appropriate bridging bus plan.

This two-stage approach effectively addresses the aforementioned challenges to some extent. For instance, this approach offers significantly improved interpretability. By separating the incident model, we enable detailed analysis of the underlying mechanisms by which incidents impact passenger flows. In contrast, an end-to-end model functions more like a “black box,” making it difficult to analyze the underlying mechanism. Furthermore, in a two-stage framework, high-quality data can be selectively utilized to calibrate the incident model, enhancing its ability to capture incident patterns. This process is more challenging for end-to-end models, as the incident data is often obscured within the normal data. Finally, our method provides flexibility, as each stage can be independently improved. For instance, if a more accurate normal OD prediction model becomes available in the future, it can be directly integrated into the first stage of our method.

In order to complete our prediction method, we will encounter the following four challenges: First, we need to develop an OD prediction model under normal conditions. Second, we must identify the causal effects of incidents on passenger flows using historical data. Third, we need to calibrate a causal effect prediction model based on the identified causal effects. Finally, we must integrate the results from the normal condition prediction model with those from the causal effect prediction model to derive the final predicted OD flows. While the first challenge is not the primary focus of this paper, we will demonstrate through experiments that any OD prediction model can be utilized in our method, regardless of its specific form. Consequently, the remaining three challenges constitute the main research questions of our study.

To identify the causal effects in historical data, the synthetic control framework is employed to estimate these effects \cite{zhang2023causalinferencedisruptionmanagement}. Besides, to differentiate actual causal effects from noise, placebo tests are utilized to assess the statistical significance of these effects. Furthermore, to calibrate the causal effect prediction model, we examine which types of effects should be used to calibrate the model and propose that only significant effects should be considered. And it is essential to balance both the quantity and quality of data when determining the threshold of significant level. Finally, in the prediction phase, we analyze which types of ODs require correction by the causal effect prediction model and conclude that only significantly affected ODs should be adjusted. Both experimental and theoretical evidence supports the importance of selecting significantly affected ODs for model calibration and passenger flow prediction.
We hope our research will provide valuable insights into the patterns of incident effects on passenger flows and offer practical tools for subway station staff to manage incidents effectively. We summarize our contributions as follows:
\begin{itemize}
    \item We proposed a two-stage OD flow prediction model for subway system during incidents. The first stage predicts
    the passenger number under normal situations and the second stage combines this prediction with the causal effect of incident to obtain the final prediction.
    \item We used synthetic control method to identify the causal effect from historical data and used placebo test to estimate its significance level.

\item  We carefully analyzed how to calibrate the causal effect prediction model and how to use it to adjust the results of normal prediction models. The importance of selecting samples that are significantly affected during the calibration and adjustment processes has been proofed both theoretically and experimentally.

\item  We conducted experiments using real-world data to demonstrate the effectiveness of our method and found some interesting patterns about how incidents affect OD passenger flows.
\end{itemize}
\section{Related Work}
\subsection{OD passenger flow prediction models}

In general, the main challenge of OD prediction is the high dimensionality \cite{Two-Stage2024}. For example, if there are 100 stations in a subway system, the number of ODs will be 10000, which is extremely challenging for traditional models. 

To address this, \cite{Cheng2021RealTimeFO,Gong2020OnlineSC} assumed that the OD matrix is of low rank and can be reduced to some much smaller matrices. Therefore, some autoregression methods could be employed for these small matrices. \cite{Two-Stage2024} assumed that the proportion of passengers traveling from one station to another, relative to the inflow number of the origin station, remains constant. As a result, the OD matrix can be directly obtained once the inflow prediction is available. \cite{Han2022ContinuousTimeAM} also used station-level embedding to estimate OD matrix and avoid the high dimension problem.

In addition, deep learning method, which is more flexible, expressive and suitable for high dimensional data \cite{lecun_deep_2015}, was used in OD prediction models. For the aspect of time -axis modeling, \cite{Toque2016ForecastingDP} used LSTM to capture temporal patterns. \cite{HybridTITS2019} employed discrete wavelet transform to capture the fluency information for OD flow in different time periods. And \cite{Ye2022CompletionAA} proposed considering day and week periods is important. Besides, attention mechanism \cite{HUANG2023ODformer} have been attempted too.

As for spatial information, \cite{Chu2020DeepMC} utilized convolutional neuron network to capture patterns in history OD matrix. \cite{Stochastic2020,Wang2019OriginDestinationMP,Makhdomi2023GNNbasedPR}  attempted graph neuron networks. And \cite{Noursalehi2021DynamicOP} used wavelet transform to extract multi-resolution features in OD matrix. \cite{Liu2021OnlineMO} introduced dual transformer to capture both temporary and spatial information and predict OD passenger flows. And \cite{Han2022ContinuousTimeAM} proposed a module to learn adaptive spatial relationships from data. In summary, despite a lot of research addressing the challenge of OD passenger flow prediction, these methods often focus on normal conditions and fail to account for the effects of incidents.
\subsection{Causality method for traffic prediction and incident effects}
Previous research has applied causality methods to predict traffic flows or traffic demand. For example, \cite{li2025causal} proposed a spatiotemporal causal intervention-based large language model approach to eliminate spurious spatial correlations and mitigate hallucination issues. \cite{li2015robust} employs Granger causality theory to select the most relevant parts from massive traffic data to build concise prediction models. \cite{khaled2024graph} uses transfer entropy to describe the causal relationship between each pair of nodes, constructs a causal adjacency matrix, and feeds it into a spatiotemporal prediction model for traffic forecasting. \cite{xu2024traffic} proposed a Multisource Causal Interpretation Graph based reasoning model to infer traffic state data for road sections without detectors. These works are highly valuable. However, their focus is conducting traffic prediction under normal situations, which differs from our research topic.

Besides, there are also some works focusing on analyzing the impact of highway incident using causal inference methods. For instance, \cite{li2024inferring} analyzed the causal effects of highway incidents on subsequent traffic conditions. Factors like weather and rush hour periods are regarded as confounding variables that simultaneously influence both the probability of incident occurrence and the resulting effects. Doubly robust learning methods are employed to eliminate the influence of these confounding factors. Similar ideas can be found in \cite{zhang2022inferring,li2016quantifying,cao2021quantification}. However, highway incidents are relatively frequent, and the datasets used in these works include thousands of incidents, enabling the calibration of a propensity score model to predict incident probability based on confounders. In contrast, subway systems are generally more stable, and our dataset contains only six incidents with significant impact. Therefore, calibrating a reliable propensity score model is highly challenging. Secondly, most subway incidents in our dataset involve faults like communication system failures or station electromechanical facility malfunctions. These may not exhibit strong confounding effects – meaning there are likely no variables that simultaneously significantly influence both the probability of an incident occurring and the impact of that incident. Therefore, even without a propensity score model might be acceptable. However, we need to address the challenge of data sparsity, where the number of treated samples (incident cases) is very small. Consequently, in this situation, we opted to use the Synthetic Control Method, which is specifically designed for scenarios with a limited number of intervention cases.

Furthermore, \cite{zhang2023causalinferencedisruptionmanagement} utilized real operational data and causal inference methods to assess the effects of subway system disruptions on station passenger flows, highlighting the spillover effects of some disruptions. However, it focuses on station-level passenger flow and only identifies the causal effects of incidents, lacking a predictive module for incident impacts. This approach is more suitable for post-event analysis rather than real-time passenger flow prediction.

\subsection{Evaluating the effects of incidents on subway systems}
Research about evaluating the effects of incident on subway systems can be broadly categorized into two main areas: network vulnerability or robustness and the impacts of incidents on passenger flows.

\subsubsection{Network vulnerability}

The majority of studies in this category utilize simulation-based methods or approaches grounded in complex network theory. These methodologies often focus on evaluating network redundancy and robustness. Complex network-based methods conceptualize the subway network as a graph, with stations as nodes and the connections between them as edges. By systematically removing nodes or edges, new graphs are generated, and researchers analyze changes in various coefficients that reflect the network's structure or passenger travel convenience \cite{Derrible2010TheCA} This allows for the assessment of incident impacts in different regions and the overall resilience of the subway network.

For instance,\cite{Zhang2011NetworkedAO} analyzed Shanghai's subway network by removing stations or intervals and evaluating changes in network efficiency, network size, average node centrality, and average edge centrality. This helped determine the vulnerability of specific intervals and assess the network's overall robustness. Similarly, \cite{Zhang2018ComparisonAO} used changes in network efficiency-related coefficients before and after node and edge deletions to evaluate the vulnerability of subway networks in several cities. Other notable studies include \cite{Qi2022ResilienceAO,Sun2018VulnerabilityAO}. \cite{Zhang2020ACI} proposed various indicators to reflect network vulnerability, using concepts from causal inference and actual data to evaluate changes in these indicators under normal and incident scenarios.

Additionally, some research incorporated network passenger flow into models to evaluate the effects of incidents on subway networks \cite{Nian2019EvaluatingTA,Liu2024AssessingMN}. Simulation-based methods were often employed to mimic passenger travel behavior under incident conditions, enabling the evaluation of passenger losses such as increased travel time and congestion costs \cite{paulsen_modelling_2018}. And some work considered the cascading failure caused by one incident \cite{Pan2024OnTD,Shen2019CascadingFA} 
 to evaluate the robustness of subway network, and found the some vulnerable stations or intervals. Furthermore, some articles proposed using simulations to quantify the spillover effects of incidents in specific intervals \cite{Yap2021QuantificationAC}.

\subsubsection{Incident effects on passenger flows}

Some studies have utilized passenger flow prediction models to evaluate the effects of incidents. For example, \cite{Silva2015PredictingTV} developed a prediction model for subway passenger flow in London under normal conditions. This model can predict expected passenger flow in the absence of an incident, so this passenger flow can serve as a counterfactual value to quantify the impact of subway incident. Based on this concept, other models designed for short-term passenger flow prediction, such as PVGCN \cite{Liu2020PhysicalVirtualCM} and MPSTN \cite{huang_leveraging_2024}, could also be adapted to assess the effects of incidents on subway stations.

Besides, \cite{Zou2024RealtimePO} created an OD prediction model using smart card data, incorporating incident status as an independent variable. This model can calculate OD values under both incident and non-incident conditions, with difference indicating the impact of subway incidents on passenger flows. Other studies have developed specific prediction models for passenger flow during incidents, comparing these outputs with predictions under normal conditions to measure the impact \cite{Xu2023ShorttermPF,Xue2021ForecastingTS,Ni2017ForecastingTS}. Simulation-based approaches have also been used to estimate passenger flow during incidents, thereby evaluating the impact at various stations \cite{Yap2020PredictingDA,Su2023SimulationBasedMF}.

Several studies have also analyzed and predicted individual passenger behaviors during incidents, which can be extended to all passengers and evaluate the change of passenger flow number due to incidents. Some of these works used survey data to understand passenger choices during incidents \cite{Rubin2005PsychologicalAB,Masud2019TravelBR}. Other research leveraged transport card data in consecutive days to identify travel behaviors under incident conditions, and compared these to normal conditions and established predictive models for passenger behavior changes during incidents \cite{Sun2016EstimatingTI}. Additionally, \cite{Mo2022InferringPR} predicted how many passengers would alter their travel behavior and how these passengers would change their behaviors during incidents.

\subsection{Summary}

In summary, existing works about passenger flow prediction under incident situation primarily apply deep learning models, often with simplistic assumptions about which passenger flows are impacted by subway incidents. These studies do not delve into how incidents affect passenger flows in detail, and the interpretability of their models is generally limited. Besides, other methods using passenger choice models to infer subway passenger flow under incident conditions typically rely on socioeconomic attributes, which may be unavailable during real-time subway operations. Studies using simulation methods for passenger flow evolution might impose overly simplistic or unrealistic rules on passenger behavior. Therefore, our paper adopts real passenger flow data, which could be complement to them.

Literature about the impact of incidents on subway systems focuses on network vulnerability, often relying on simulations or complex network methods that may not fully reflect real-world subway operations. Besides, their research questions are different compared with the questions proposed in this work. Some studies evaluate the impact of incidents on passenger flow using predictive models, comparing predicted normal conditions with actual passenger flow during incidents. However, these studies lack causality analysis. Additionally, they can only calculate these differences after incidents ended. 
\section{Method}
Our method consists of three parts: causal effect estimation, causal effect prediction model calibration and normal prediction adjustment. We will elaborate them in the following part.
\subsection{Causal effect estimation}

We use the synthetic control method \cite{Abadie2010SyntheticCM} to calculate counterfactual values and obtain causal effects. This methods have been used in transportation section and many researchers used them to evaluate policy effects. For example, \cite{yang2022assessing,dai2021improving} assess the effects of subway price policy on passenger flows and \cite{wallimann2023price} evaluate the effects of new railway lines on congestion. Following the algorithm proposed in \cite{Abadie2010SyntheticCM}, we designed our approach to estimate the causal effects. At first, we will give some notations.

Assuming we have a dataset containing $n$ days, with each day divided into m time intervals, let $x_{i,j}$ represent the passenger flow for a specific OD on day $i$ during time interval $j$. Here, $x_{i,j}^0$ denotes the passenger flow during normal conditions, while $x_{i,j}^1$  indicates the flow after an incident occurs. For simplicity, we assume only one incident occurs in the $k$-th time interval of day $l$ (our method can be easily generalized to multiple incident situation). Thus, the observed data for a specific OD can be expressed as:
\begin{equation}
\begin{array}{cccccccc}
    x_{1,1}^{0}&x_{1,2}^{0}&\cdots&x_{1,k-1}^{0}&x_{1,k}^{0}&x_{l,k+1}^{0}&\cdots&x_{1,m}^{0}\\
    x_{2,1}^{0}&x_{2,2}^{0}&\cdots&x_{2,k-1}^{0}&x_{2,k}^{0}&x_{2,k+1}^{0}&\cdots&x_{2,m}^{0}\\
    \vdots&\vdots&\ddots&\vdots&\vdots&\vdots&\ddots&\vdots \\
    x_{l-1,1}^{0}&x_{l-1,2}^{0}&\cdots&x_{l-1,k-1}^{0}&x_{l-1,k}^{0}&x_{l-1,k+1}^{0}&\cdots&x_{l-1,m}^{0}\\
    x_{l,1}^{0}&x_{l,2}^{0}&\cdots&x_{l,k-1}^{0}&\bm{x_{l,k}^{1}}&\bm{x_{l,k+1}^{1}}&\cdots&\bm{x_{l,m}^{1}}\\
    x_{l+1,1}^{0}&x_{l+1,2}^{0}&\cdots&x_{l+1,k-1}^{0}&x_{l+1,k}^{0}&x_{l+1,k+1}^{0}&\cdots&x_{l+1,m}^{0}\\
    \vdots&\vdots&\ddots&\vdots&\vdots&\vdots&\ddots&\vdots \\
    x_{n,1}^{0}&x_{n,2}^{0}&\cdots&x_{n,k-1}^{0}&x_{n,k}^{0}&x_{n,k+1}^{0}&\cdots&x_{n,m}^{0}\\
\end{array}
\label{matrix data}
\end{equation}
And we want to estimate: $x_{l,k}^1-x_{l,k}^0$, $x_{l,k+1}^1-x_{l,k+1}^0$,…,. It is important to note that $x_{l,k}^1$ is observed, while $x_{l,k}^0$ is not. The latter term represents a  \textit{counterfactual value} in the context of causal inference. 

Additionally, let $a_{i,j}$ denote the influencing factors for $x_{i,j}$, such as whether day $i$ is rainy, whether it is a weekend, and passenger flow numbers from previous intervals. We assume the dimension of $a_{i,j}$ is $d$.

\begin{figure}
    \centering
    \includegraphics[width=\linewidth]{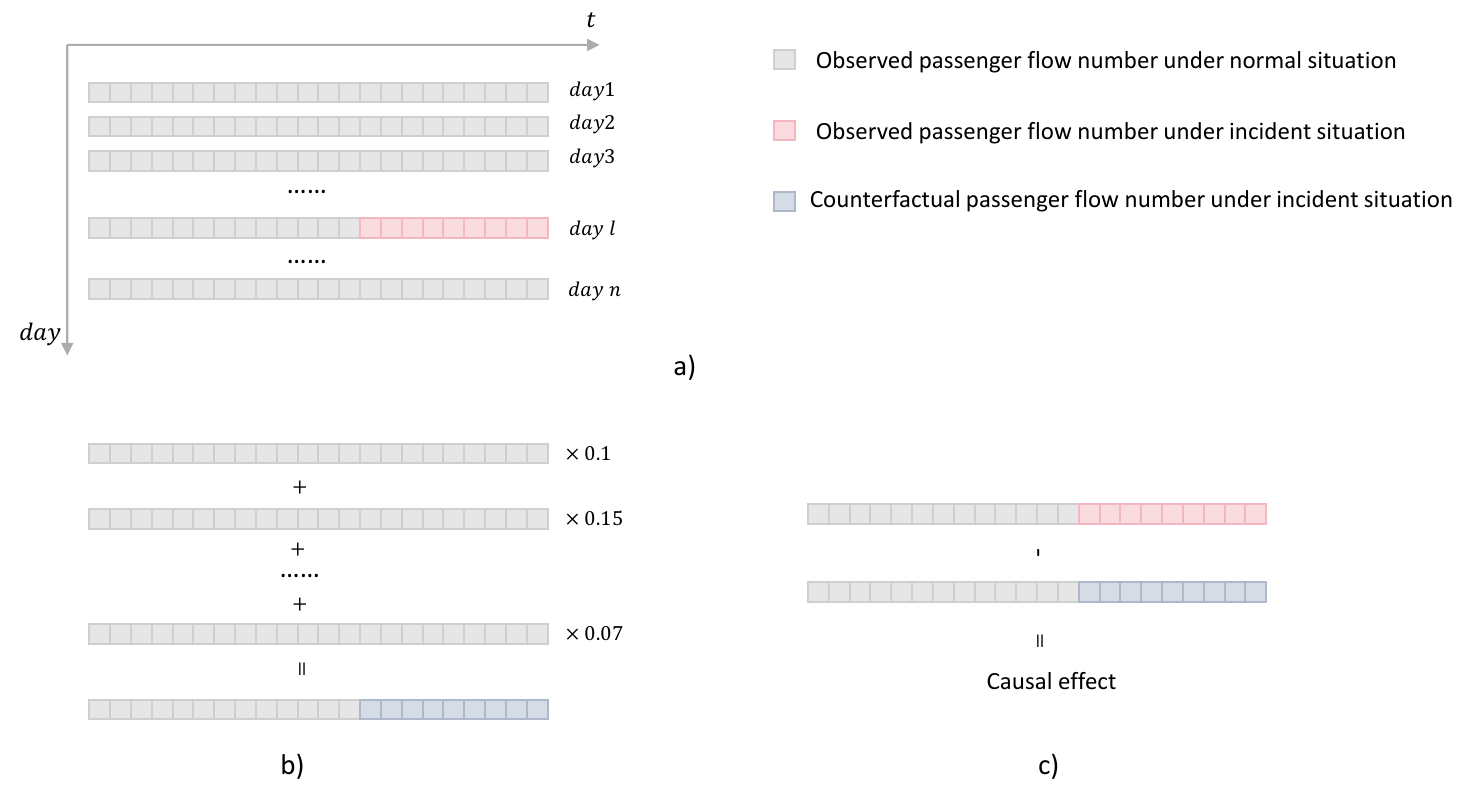}
    \caption{Process of causal effect evaluation: a) Observed data. b) Using observed data in normal cases to synthesize counterfactual data. c) Calculating causal effect.}
    \label{fig:data_des}
\end{figure}

The process for synthetic control can be summarized as: \textit{synthetic the unobserved $x^0$ using observed $x^0$}, as illustrated in Figure \ref{fig:data_des}, and we will explain this process using the calculation of $x_{l,k}^0$ as an example.

In our setting, $x_{i,k}^0$ is observed for $i\in 1,2,…,l-1,l+1,…,n$. We estimate $x_{l,k}^0$ using the following formulation:
\begin{equation}
    \hat{x}_{l,k}^{0} = \sum_{i \in \{1,2,\ldots,l-1,l+1,\ldots,n\}} w_i \times x_{i,k}^{0}
\end{equation}
where each $w_i$ is in $[0,1]$ and $\sum_i{w_i}=1$. This means that if there was no incident on day $l$, the passenger flow on that day would approximate a weighted-sum of passenger flow numbers on other no-incident days. In other words, we synthesize counterfactual data using observed data, with "synthetic" referring to the weighted sum.
For convenience, we use W to signify the vector composed of all $w_i$'s: $W = (w_1,…w_{l-1},w_{l+1},…,w_n )\in R^{n-1}$. The challenge then becomes determining $w_i$. Recall that a $d-$dimensional vector $a_{i,j}$ is defined to represent the influence factors of $x_{i,j}$. Therefore, the target of determining $W$ is to make the variable $a$ of the synthetic sample more similar to the $a$ in day $l$. Let $A$ be the matrix of all $a_{i,j}$'s where  $i\in \{1,2,…,l-1,l+1,…,n\}$. Thus, $A\in R^{n-1,d}$. Then $W$ can be determined as:
\begin{equation}
    W = \mathop{\arg\min}_{W} \|a_{l,k} - WA\| = \sqrt{(a_{l,k} - WA)(a_{l,k} - WA)^{T}}
    \label{eq:w1}
\end{equation}
But the importance of each dimension in $a$ may not be equal. As a result, \cite{Abadie2021UsingSC} propose to use $V-$norm instead of $l_2$ norm and we follow this setting in our method. So $W$ is determined by:
\begin{equation}
    W = \mathop{\arg\min}_{W} \|a_{l,k} - WA\|_{V} = \sqrt{(a_{l,k} - WA)V(a_{l,k} - WA)^{T}}
    \label{eq:w}
\end{equation}
where $V$ is a $d\times d$ positive diagonal matrix. Equation \ref{eq:w} means the influence factors of synthetic counterfactual data, i.e., $WA$, are similar to the influence factors of actual counterfactual data $a_{l,k}$. The matrix $V$ represents the importance of each dimension in a. According to \cite{Abadie2021UsingSC}, $V$ can be chosen to minimize the prediction error in previous $t$ time intervals, as the following Equation \ref{eq:v} shows:
\begin{equation}
    V = \mathop{\arg\min}_{V} \|XW(V) - X_0\|
    \label{eq:v}
\end{equation}
where $X_0$ is a $t-$dimensional vector, $X_0  = (x_{l,k-t}^0,x_{l,k-t+1}^0,…,x_{l,k-1}^0)$, and $X$ is a matrix with shape $t\times (n-1)$, the $i$-th column of $X$ representing the passenger flow number in time intervals $k-t$ to $k-1$ of day $i$. $W(V)$ is the function mapping $V$ to $W$, as defined by Equation \ref{eq:w}. For more details, interested readers can refer to \cite{Abadie2021UsingSC}. Finally, the causal effect can be estimated as $x_{l,k}^{1} - \hat{x}_{l,k}^{0}$.

Some readers may question that why we do not choose to fit a model form $a$ to $x^0$ directly but use such a complicated process. This is because directly fitting model could meet the problem of extrapolation and results in inferior performance, this phenomenon has been discussed by \cite{Abadie2014ComparativePA} and interested readers can refer to it.
\subsection{Causal effect prediction model}
Even though we can identify the causal effects of incidents on OD flows, these effects are determined only after the incidents have ended and cannot be directly applied to the actual operation of the subway system. Therefore, we need a prediction model to predict the causal effects of incidents on passenger flows immediately after an incident occurs.
\subsubsection{Feature selection}
First, it is necessary to identify the factors influencing the causal effects. On one hand, the severity of an incident is believed to impact these causal effects. It is reasonable to assume that more severe incidents have a greater potential to affect a larger number of ODs and result in more significant consequences. We have selected six primary indicators to quantify severity: the maximum delay time caused by the incident, the number of trains delayed by five minutes or more, the number of canceled trains, the number of trains that have cleared all passengers, the number of affected stations, and the overall duration of the incident.
\begin{figure}
    \centering
    \includegraphics[width=0.9\linewidth]{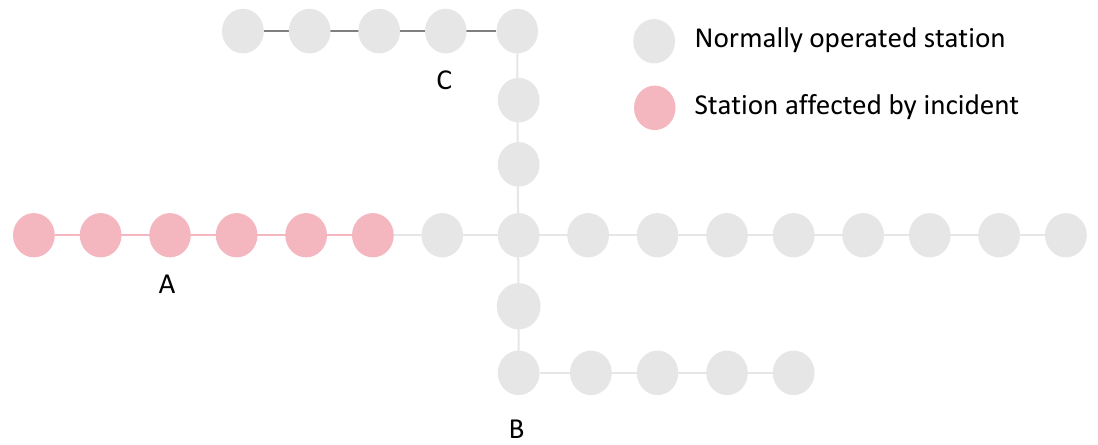}
    \caption{An example of incident}
    \label{fig:incident_example}
\end{figure}

On the other hand, the relationship between the ODs and the incident, in both spatial and temporal dimensions, is also considered to influence the causal effects. In the spatial domain, we consider several indicators: the distance from the origin station to the incident interval (where distance is defined as the number of stations on the shortest path, and this definition applies to subsequent distances as well), the distance from the destination station to the incident interval, and the proportion of the shortest path between the origin and destination that overlaps with the incident interval. These indicators reflect the spatial relationship between the ODs and the incident. In the temporal domain, we select the time difference between the OD time and the occurrence of the incident, the time difference between the OD time and the end of the incident, and whether the OD time falls within the duration of the incident. Finally, if the OD originally experiences a higher passenger flow, the causal effect may be more pronounced. Consequently, the counterfactual flow $x_0$ is also selected.

It is important to note that certain indicators, such as the delay time caused by an incident, the number of trains delayed by five minutes or more and the overall duration of the incident, can only be obtained after the incidents have concluded. Consequently, these indicators cannot be accessed when predicting causal effects. However, we believe this is not a significant issue, as experienced staff in the subway system can generally extrapolate these indicators based on the type or other characteristics of the incident. Additionally, the severity of the incident must be considered when predicting its causal effects; therefore, we have no choice but to include these indicators. In future work, features related to the types of incidents and the degree of damage to facilities could be considered as indicators of incident severity, and these features are avaliable immediately after an incident occurs. A description of the selected features is provided in Table \ref{tab:Feature Description}.
\begin{table}[!h]
    \centering
    \begin{tabular}{cp{8cm}c}
        \toprule
         Feature & Meaning &Unit \\
        \midrule
        
        duration & the duration of the incident & minute \\
        max\_delay & the delay time caused by the incident & minute \\
        delay\_5\_num & the number of train trips delayed by five minutes or more&  \\
        evacuate\_num & the number of trains that required evacuation &  \\
        cancel\_num &  the number of canceled trains &  \\
        influence\_station\_num & the number of stations within the affected interval &  \\
        distance\_d\ & the shortest distance from the OD's destination station to the incident interval &  \\
        distance\_o\ & the shortest distance from the origin station to the incident interval &  \\
        proportion & the proportion of the shortest path between the origin and destination stations that overlaps with the affected interval & \\
        time\_diff\_to\_start & the time difference between the OD time and the incident occurrence time & minute \\
        time\_diff\_to\_end & the time difference between the OD time and the end of the incident & minute \\
        is\_in\_incident & whether the OD time falls within the incident's duration &  \\
        $x_0$&Predicted counterfactual passenger flow number& \\
        \bottomrule
    \end{tabular}
    \caption{Feature Description}
    \label{tab:Feature Description}
    
\end{table}

\subsubsection{Sample selection}

\begin{figure}
    \centering
    \includegraphics[width=\linewidth]{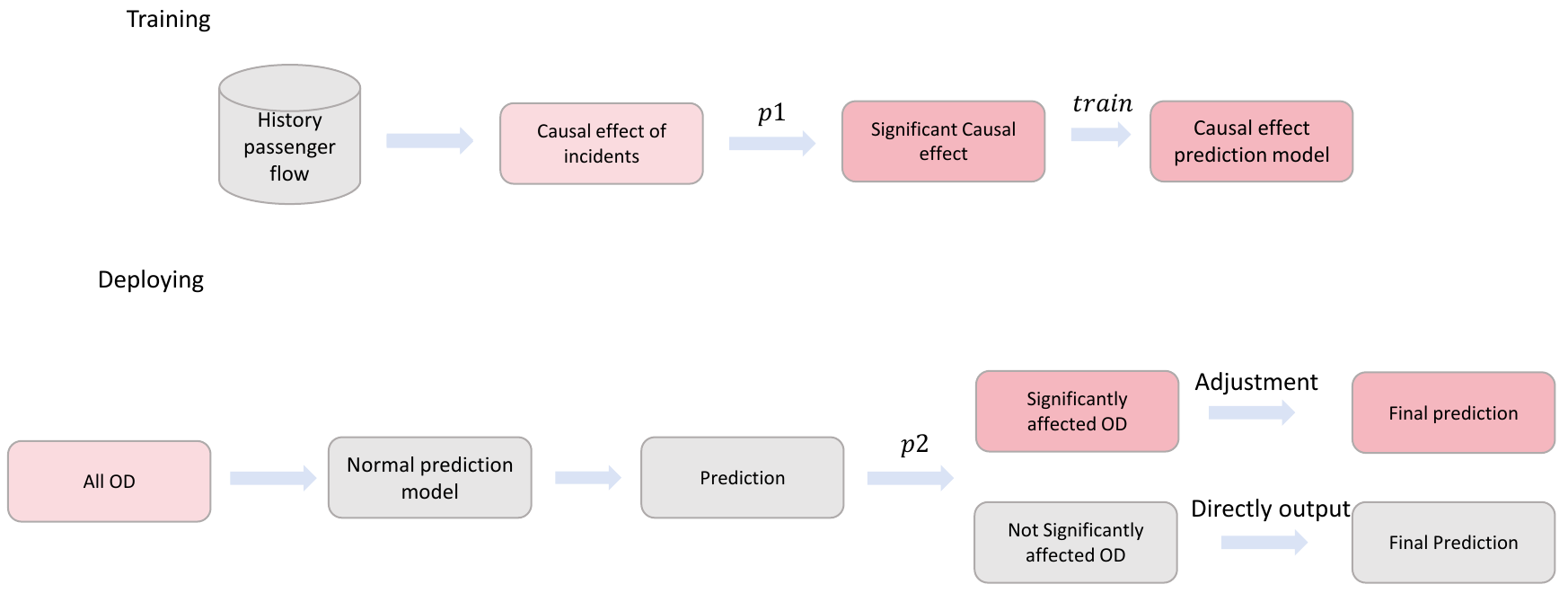}
    \caption{The details of training and deploying causal effect prediction model}
    \label{fig:wolkflow_detail}
\end{figure}

According to the previous section, we can calculate the causal effects for all OD pairs and all time intervals for an incident. However, these differences are not solely attributable to the incident; some may be influenced by random effects. For example, an incident situation is show in Figure \ref{fig:incident_example}, and this incident may clearly impact the passenger flow from station A to station B, but the effect on the passenger flow from station C to station B may be less apparent. Although the difference between the counterfactual and observed passenger flow from station C to station B can indeed be calculated, it is essential to determine whether this difference is due to random effects or the incident itself. Therefore, it is necessary to select the significantly affected passenger flows, as illustrated in the upper part of Figure \ref{fig:wolkflow_detail}.

We provide a theoretical explanation of why selecting a truly affected sample to calibrate a causal prediction model is important. First, let us consider the following linear model.

\begin{definition}
If the influence factors about the causal effects is noted as $x$, the causal effect is denated as $y$, assume $y$ and $x$ obeys the following rule:

$y=\epsilon_1$ with probability $1-p$;

$y=\epsilon_1+(\beta^T x+\epsilon_2)$ with probability $p$

Where $\epsilon_1$ follows normal distribution $N(0,\sigma_1^2)$ and $\epsilon_2$ follows normal distribution $N(0,\sigma_2^2)$, $\beta$ is the parameter and $p$ is a constant between 0 and 1. 
\label{noisy-sample}
\end{definition}

This model indicates that in our dataset, only a proportion $p$ of the samples are truly affected samples, while the remaining samples, despite the ability to derive causal effects, are influenced by randomness. Furthermore, the causal effect is a linear function of the influencing factors. We then present a theorem to describe the consequences of fitting a linear model with all samples.

\begin{theorem}
    If there are $n$ samples obeying the above model and we directly fit a model with these samples, resulting in the fitted parameter $\hat{\beta}$. Then, the square distance between true parameter and the fitted parameter is:
    \begin{equation}
        E(\beta-\hat{\beta} )^2=(1-p)^2\ |\beta|^2+c\frac{p\sigma_2^2+\sigma_1^2}{n} 
    \end{equation}

    Where $c$ is a constant.
    \label{theo:3.1}
\end{theorem}

It can be found that the distance is bound by two terms, the first term is $\left(1-p\right)\left|\beta\right|^2$, which means the larger the $p$ is, the smaller the loss will be. This law is consistent with normal intuition because if more samples in the dataset are actually affected by incidents, the fitting results will be better. The last term $\frac{p\sigma_2^2+\sigma_1^2}{n}$ is related to sample size $n$, which means that the more samples are in the dataset, the better the fitting result will be. This law is also intuitive. In summary, the theorem means that we need to consider both the sample quantity ($n$) and the sample quality ($p$) when selecting which sample should be used to fit the prediction model. As a result, if we just use all OD records after incidents, then the sample quantity $n$ could be large but sample quality $p$ might be small. Therefore, if we can propose a method to judge which causal effect is more likely to be true effect and which is more likely to be randomness, then we can select these true effects and control the sample quality $p$ to be larger and obtain a better fitting result.

Even though we provide only a theoretical analysis of the effect of sample selection on linear models, our experiments indicate that this principle also applies to complex machine learning models.
\subsubsection{Significant test}
Judging a causal effect is truly because of incident or because of randomness is a classical significant test problem. And we will elaborate it in the following part.

First, our null hypothesis is:
\begin{equation}
    H_0: x_{l,k}^{0} = x_{l,k}^{1}
    \label{eq:null_hype}
\end{equation}
This null hypothesis means the actual OD flow under incident $x_{i,k}^1$ is the same as the counterfactuals OD flow in normal situation $x_{i,k}^0$. Therefore, the calculated causal effect, i.e., $x_{i,k}^1-{\hat{x}}_{i,k}^0$, is because of randomness. We use a placebo test to construct the test statistic \cite{Abadie2010SyntheticCM} and determine whether to reject the null hypothesis. If the null hypothesis is rejected, then we can regard this causal effect is different from zero significantly. The placebo test is widely used, distribution-free and similar to the permutation test. We will elaborate it as follows.

Recall the method from the causal effect estimation section, we use $(a_{i,k}, x_{i,k})$ for $i\in\ \{1,2, \ldots, l-1, l+1, \ldots, n\}$ and $a_{l,k}$ to infer $x_{l,k}^0$. Let $f$ represent this process. Besides, let $S$ denote the set $\{(a_{i,k},\ x_{i,k})\mid\ i\in \{1,\ 2, \ldots, n\} \}$, and $S_{-i}$ denote the set obtained by removing $(a_{i,k},\ x_{i,k})$ from $S$. Thus, we have:
\begin{equation}
    \hat{x}_{l,k}^{0} = f(S_{-l}, a_{l,k})
\end{equation}
Then we define $e_i = |{\hat{x}}_{i,k}^0\ -x_{i,k}| = |f(S_{-i},a_{i,k})-x_{i,k}|$, where $x_{i,k}$ represents the observed passenger flow. Specifically, $x_{i,k}=x_{i,k}^1$ when $i=l$, and $x_{i,k}=x_{i,k}^0$ on other days. Therefore, $e_1, e_2, \ldots, e_n$ can be calculated according to our definition.

Intuitively, if the null hypothesis is true, which means $x_{l,k}^0\ =x_{l,k}^1$, then all the $(a_{i,k}, x_{i,k})$ follow the same rule since all $x_{i,k}$ equal to $x_{l,k}^0$. Consequently, each error term $e_i$ will follow the same distribution. The order of $e_l$ is equally likely to be in any position from 1 to $n$. In another word, if $e_l$ is extremely large, for example, the largest among all $e_i$, it indicates a low probability event happens. Thus, the order of $e_l$ can be viewed as a test statistic to control the type 1 error in the significance test, i.e., to control the significance level.

Let $order(e_l)$ denote the order of $e_l$ among $e_1, e_2,\ldots, e_n$. For example, $order(e_l) = n$ means $e_l$ is the largest element. Then we can use $order(e_l)$ to construct p-value as: 
\begin{equation}
    p=1-\frac{order\left(e_l\right)}{n}
\end{equation}
If the significance level we want to control is $\alpha $ (which is $p1$ in Figure \ref{fig:wolkflow_detail}) and $p\le\alpha$, we will reject $H_0$ and conclude that the incident significantly influenced the passenger flow. 
\subsection{Prediction adjustment}
In practical forecasting tasks, we first utilize a conventional passenger flow prediction model to obtain passenger flow under normal conditions. Subsequently, we utilize a causal effect prediction model to derive the predicted causal effects of specific incident. These causal effects are then used to adjust the passenger flow predictions from the normal model, resulting in passenger flow forecasts under incident conditions. In this section, we concentrate on how to use the causal effect prediction model to correct normal predictions. This approach may seem somewhat peculiar, as one might assume that adjustments could simply be summed together. However, our research underscores the importance of determining which normal predictions should be adjusted.

Taking the example illustrated in Figure \ref{fig:incident_example}, it is evident that the passenger flow from station A to station B is likely to be significantly affected by an incident. Therefore, we need to add the causal effect to the predicted passenger flow for the OD from station A to station B. However, the passenger flow from station C to station B raises questions. Should we also adjust the passenger flow for this OD? This remains uncertain, and thus we need to ascertain which passenger flows require adjustment.

We use the following strategy to adjust OD flow prediction: If there are many ODs and these ODs are affected by an incident with different probability according to relationship between these ODs and the incident. Then we note the probability of a specific OD to be affected by the incident is $p$. Then the strategy is: First determining a threshold $P$ (which is $1-p2$ in Figure \ref{fig:wolkflow_detail}), for each OD, if the probability of the OD to be affected by the incident, i.e., $p$, is greater than $P$, we will use causal effect prediction model to adjust it, otherwise we will just regard the prediction of normal model as the final prediction value, as shown in Figure \ref{fig:wolkflow_detail}. Then we can obtain the risk of this strategy.

\begin{theorem}
If the influence factors about the causal effects is noted as $x$, the causal effect is donated as y, assume y and x obeys the following rule:

$y=\epsilon_1 \ with\  probability\  1-p$;

$y=\epsilon_1+(f(x)+\epsilon_2)\ with\ probability\ p$

Where $\epsilon_1$ follows normal distribution $N(0,\sigma_1^2)$ and $\epsilon_2$ follows normal distribution $N(0,\sigma_2^2)$, $\beta$ is the parameter and $p$ is a constant between 0 and 1. 
If there are a lot of samples $\left(x_i,p_i\right)$ and a prediction model $\hat{y}=\hat{f}(x)$.
Then we can get the risk of the above strategy.
\begin{equation}
E\left(y-\hat{y}\right)^2=\frac{1}{2}P^2(Ef\left(x\right)^2+E\hat{f}\left(x\right)^2-E\left(\hat{f}\left(x\right)-f\left(x\right)\right)^2)-PE\hat{f}\left(x\right)^2+c
\end{equation}
Where $c$ is a constant. and the expectation is taken over $x$ and $\epsilon_1,\epsilon_2$.
\label{thro:adjust}
\end{theorem}

The risk is a quadratic function of $P$, then to obtain the minimum risk, we should set $P$ as:
\begin{equation}
    P=\frac{E\hat{f}\left(x\right)^2}{Ef\left(x\right)^2+E\hat{f}\left(x\right)^2-E\left(f\left(x\right)-\hat{f}\left(x\right)\right)^2}  
    \label{eq:P}
\end{equation}

Some detailed analysis of this theorem is given as follows. First, when $\hat{f}(x)=f(x)$, which means the causal effect prediction model is completely correct, then we have $P=1/2$. This strategy means that if we know the probability for an OD to be affected by an incident is great than 0.5, we will use the causal effect prediction model to adjust this OD passenger flow, otherwise, we will not adjust it. Besides, when $E\left(f\left(x\right)-\hat{f}\left(x\right)\right)^2$is smaller, in other words, the mean square error of causal effect prediction model is smaller, the best $P$ could be smaller. This rule means more samples can be adjusted if the model is more accurate and it also conforms intuition. Finally, when the magnitude of causal effect, i.e. $Ef\left(x\right)^2$, is greater, the $P$ should be set to a smaller value, which means more samples should be adjusted.

Moreover, the probability that a specific OD pair is affected by an incident ($p_i$) is not directly observable. However, we can utilize historically derived p-values (calculated in the significance test section) along with relevant influence factors to develop a predictive model for $p_i$. In our actual experimental implementation, we employ a random forest model and use identical predictors in the causal effect prediction model to predict the significance level of causal effects. As a result, the workflow of training and deploying the causal effect prediction model could be shown in Figure \ref{fig:wolkflow_detail}, and $p1,p2$ in the figure are thresholds used to select training samples and samples need correction. 
\section{Experiments}
\subsection{Data description}
The data utilized in our study is derived from the Shanghai subway system, one of the largest and most complex subway networks in China, which comprises 20 lines and 508 stations.

The first part comprises OD data from the Shanghai subway system. The timing of ODs is determined by the exit time, with a temporal granularity of thirty minutes. The data covers the period from January 1, 2024, to March 24, 2024. We excluded data from the New Year's holiday and the Chinese New Year period because their unique passenger flow patterns differ significantly from those of regular weekdays.

Besides, it should be emphasized that exit-time OD cannot be used to generate bridge bus directly. In practice, exit-time OD needs to be subtracted from OD travel time to transform into entry-time OD. Moreover, if entry-time OD data become accessible in the future, developing models for entry-time OD should be attempted.

The second part of our dataset includes incident records from the Shanghai subway system. These records provide a detailed description of each incident, including the start time, end time, maximum delay caused by the incident, the number of canceled trains, and other relevant information.

Most incidents in our dataset had minimal impact, resulting in no or insignificant delays. Consequently, we only selected those incidents that caused delays exceeding five minutes. Additionally, we excluded incidents that occurred during the early morning hours, such as those involving the first train of the day, as passenger flow during these times is inherently low, rendering the analysis less meaningful. Ultimately, we identified six eligible incidents, with the last incident in our dataset occurring on March 14th. Our objective is to utilize the data prior to March 14th to build models that predict passenger flow on that day. It is noted that the incident began at 7:15 a.m. and ended at 9:24 a.m.; therefore, our focus is to predict passenger flows between 7:00 a.m. and 12:30 p.m. Given that the incident occurred in suburban sections, OD pairs whose origin and destination stations are not located on the incident line are unlikely to be affected. Therefore, our analysis concentrates on ODs with at least one station (either the origin or destination) located on the incident line.
\subsection{Causal effect identification results}
The details in causal effect identification are first explained. As previously mentioned in Equation \ref{eq:w} of Method section, we need to consider several influencing factors $a$ for an OD. In this paper, we use weather conditions (whether it is sunny), whether it is a weekend, and historical passenger flow volume from the previous two time periods as $a$. Similarly, when we determine V using Equation \ref{eq:v}, we also rely on historical data from the previous two time periods to construct $X_0$ and $X$. The significance level used in our analysis is 0.05. 

The effect of an incident typically increases with its onset and gradually dissipates as the incident ends. Consequently, we analyze all OD pairs from the beginning of an incident until three hours after its end. Using the model proposed in the Method section, we identify the causal effects of each subway incident and conduct significance tests on these effects. However, due to space limitations, we only provide a detailed report on the causal effect identification results for the incident in Marth 11th. The details of this incident are as follows:

Starting at 7:45 on the Marth 11th, a vehicle malfunction occurred on Line 11, and the malfunctioning vehicle was arranged to proceed slowly. From 7:51 to 8:43, temporary passenger flow control measures were implemented at Malu, Chenxiang Highway, Nanxiang, Taopu Xincun, Qilianshan Road, Liziyuan, and Fengqiao Road stations.

\begin{figure}
    \centering
    \includegraphics[width=\linewidth]{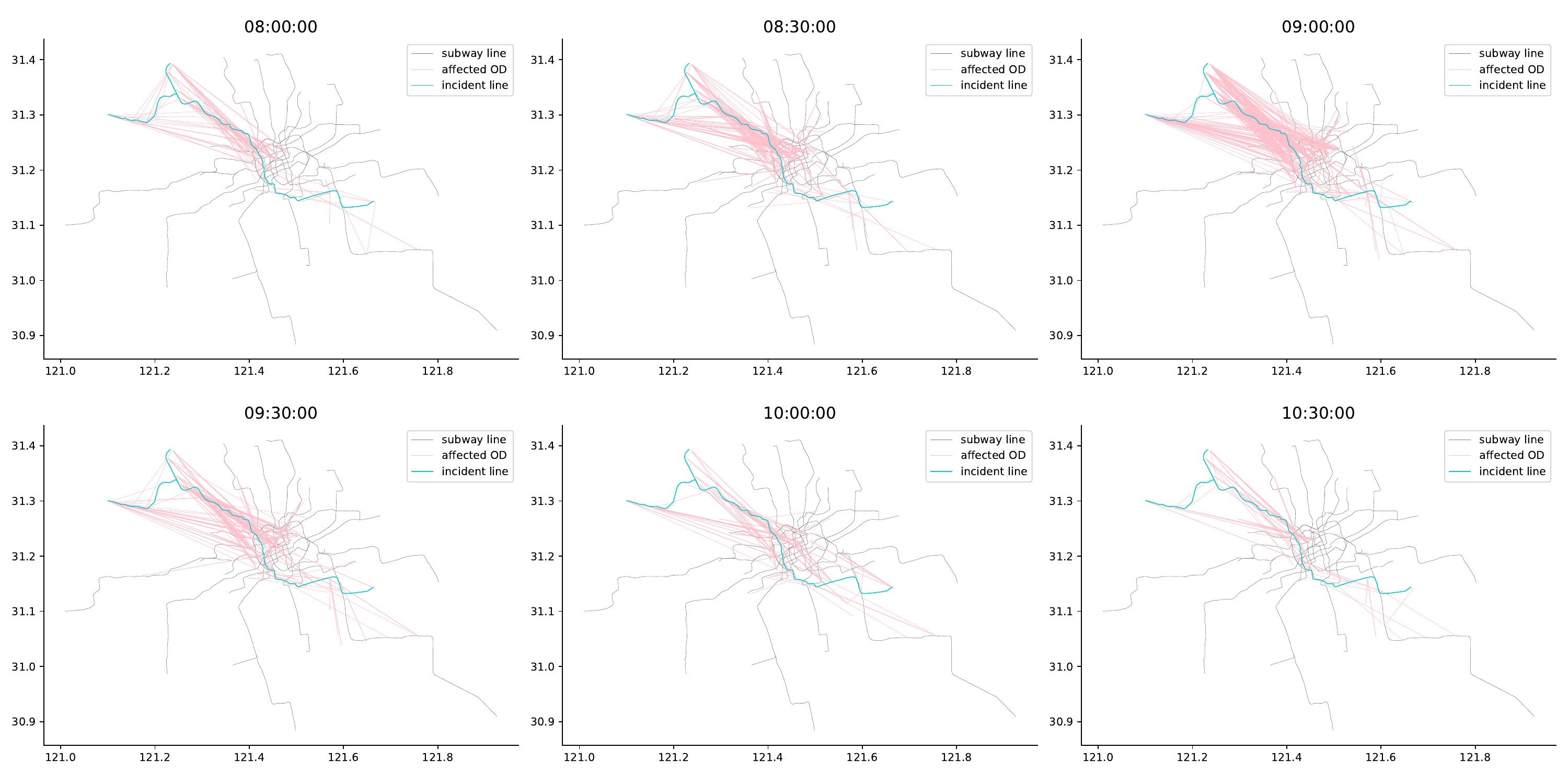}
    \caption{Significantly influenced ODs in different time intervals}
    \label{fig:affected_od}
\end{figure}

Figure \ref{fig:affected_od} illustrates the significantly affected ODs identified during the six time periods from 8:00 a.m. to 10:30 a.m.. In the figure, the blue line represents the subway line where the incident occurred, the black lines denote regular subway lines, and the red lines indicate the ODs that were significantly impacted by the incident.

It should be noted that the incident occurred at 7:45 a.m. and ended at approximately 8:45 a.m.. As illustrated in Figure \ref{fig:affected_od}, the number of affected ODs initially increased and peaked between 8:30 and 9:00 a.m., before gradually declining. By 10:00 and 10:30 a.m., only a few ODs remained affected. Additionally, at the onset of the incident, relatively few ODs were impacted, as the effects had not yet spread widely.

Furthermore, the affected stations are primarily located in the northern part of Shanghai. Figure \ref{fig:affected_od} confirms that the ODs affected by this incident are predominantly situated in the northern region, which is consistent with the actual situation.

\begin{figure}
    \centering
    \includegraphics[width=\linewidth]{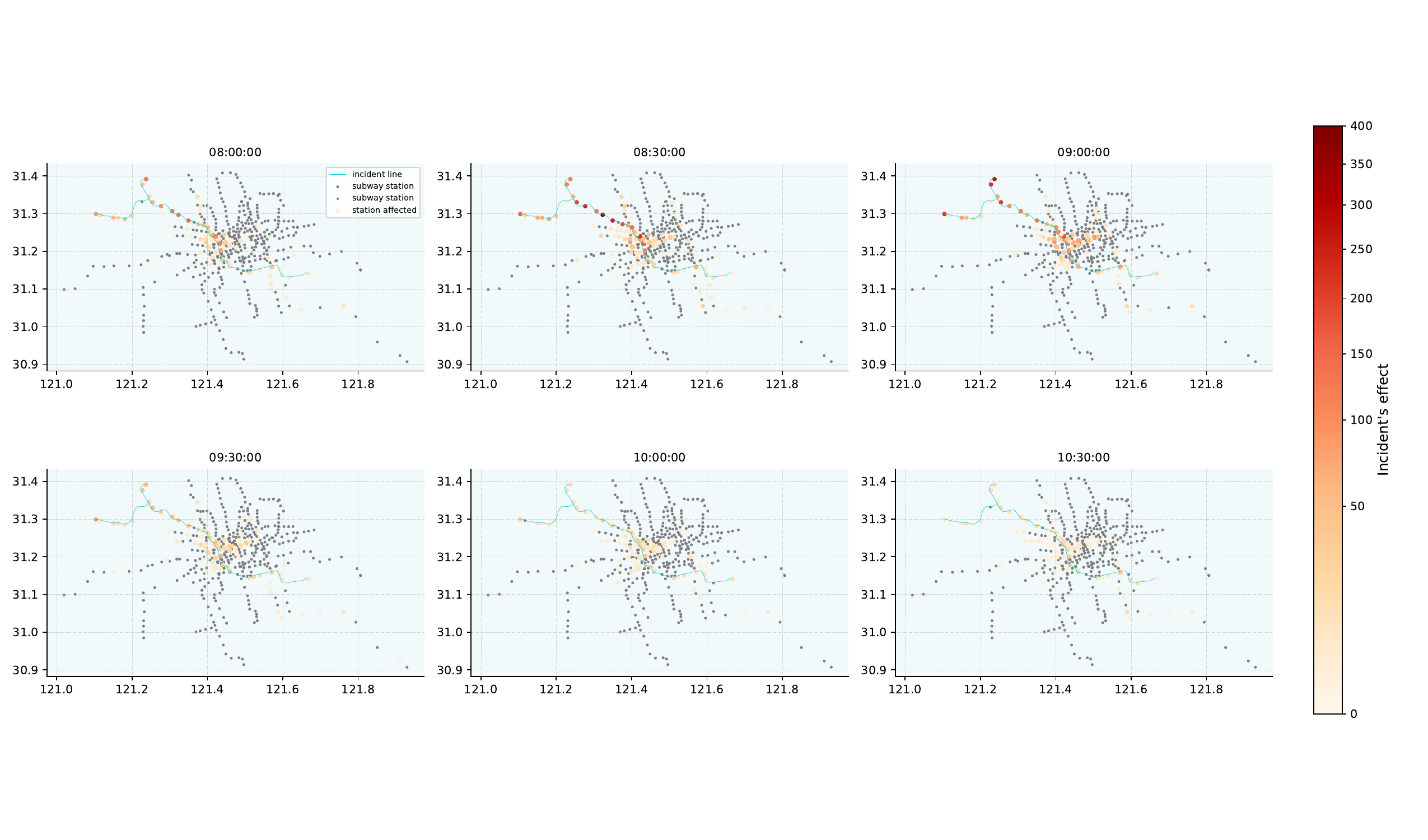}
    \caption{Incident’s effects on stations in different time}
    \label{fig:subway_disruption}
\end{figure}
Besides, Figure \ref{fig:subway_disruption} illustrates the stations affected over various time periods and the corresponding levels of impact. The impact level of each station is determined by summing the absolute values of causal effects across all ODs related to this station. Black points represent unaffected stations, while the colors of affected stations range from yellow to red, with redder points indicating higher levels of impact. The blue line denotes the incident subway route. It can be concluded that between 8:00 and 9:00, a greater number of stations were affected, exhibiting stronger impacts. As time progressed, both the number of affected stations and the levels of impact decreased.
\begin{figure}
    \centering
    \includegraphics[width=\linewidth]{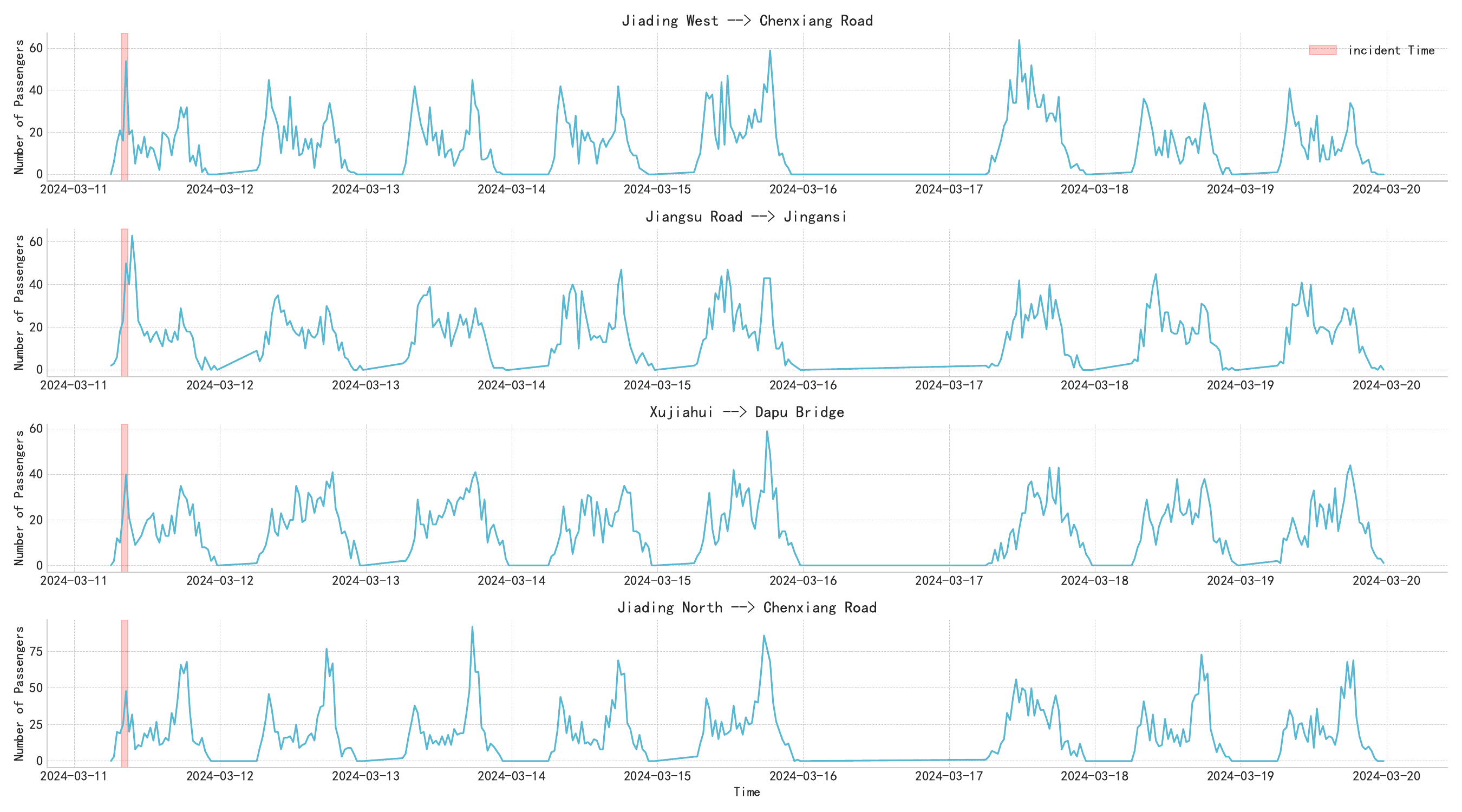}
    \caption{Passenger flow numbers of four significantly influenced ODs}
    \label{fig:od_plot}
\end{figure}

For further analysis, four significantly affected ODs are selected: Jiading West to Chenxiang Road, Jiangsu Road to Jing'an Temple, Xujiahui to Dapu Bridge, and Jiading North to Nanxiang. The passenger flow numbers for these four ODs over an eight-day period, from March 11th to March 20th are illustrated in Figure \ref{fig:od_plot}.

First, it is important to note that March 11th is a Monday, and the following Monday is March 18th. Additionally, there is no data for March 16th due to missing data in the original dataset. Each subplot in Figure \ref{fig:od_plot} illustrates the changing trend of an OD, with the red box on March 11th indicating the time of the incident.

For these four ODs, the patterns observed during the incident period differ significantly from those on other days. For example, the top subplot, which illustrates the OD from Jiading West to Chenxiang Road, indicates that the OD value within the red box is considerably higher compared to other weekdays, such as March 12th, March 13th, and the following Monday, March 18th. Similar patterns can be observed in the other three ODs. This phenomenon may be attributed to the primary consequence of the incident: delays occurring around 8:00 a.m.. These delays could lead passengers who would typically take the subway at 8:00 a.m. to board later, while those who planned to travel after 8:00 a.m. continued with their original schedules. This overlap of passengers resulted in a heightened peak during the morning rush hour.
\subsection{Passenger flow prediction results}
In this section, we conducted experiments to determine whether the proposed two-stage approach enhances the accuracy of passenger flow predictions under incident conditions. We began by selecting baseline prediction models under normal situations. These models include three traditional machine learning algorithms: Linear Regression (LR), Random Forest (RF), and Gradient Boosted Decision Trees (GBDT). 

We employed 5-fold cross-validated grid search to determine optimal parameters for RF and GBDT models. For the RF implementation, we evaluated n\_estimators across [100, 300, 500], max\_depth across [5, 10, 15] and min\_samples\_split across [2, 5, 10]. For the GBDT framework, we optimized n\_estimators within [200, 500, 800], learning\_rate across [0.01, 0.05, 0.1], max\_depth in [3, 5, 7], and subsample ratio between [0.8,0.9, 1.0].

We selected two matrix-based models: ST-ResNet \cite{Zhang_Zheng_Qi_2017} and UNet \cite{ronneberger2015u}, which are widely utilized  for matrix data, and GEML \cite{Wang2019OriginDestinationMP}, a model specifically designed for Ride-hailing OD prediction.

We also conducted experiments with different causal effect prediction models. Specifically, we selected three models: Linear Regression, Random Forest, and Gradient Boosted Decision Trees. Following \cite{Zou2024RealtimePO}, we omitted the ODs with passenger flow less than 2. Besides, MAE (mean absolute err), RMSE (root mean square error) and MAPE (mean absolute percentage error) are used to evaluate the prediction accuracy. These metrics are calculated as follows:
\begin{equation}
MAE=\frac{1}{n}\sum{|\hat{y}-y|}
\end{equation}
\begin{equation}
RMSE=\sqrt{\frac{\sum{(\hat{y}-y)^2}}{n}}
\end{equation}
\begin{equation}
MAPE=\frac{1}{n}\sum{\frac{{|\hat{y}-y}|}{y}}
\end{equation}

Where $\hat{y},y$ are the true and predicted OD flows, and $n$ is the total number OD flows. The results of experiments are reported in Table \ref{tab:result}. 

\renewcommand{\arraystretch}{1.2}
\newgeometry{top=1in, bottom=1in, left=0.5in, right=0.5in}

\begin{landscape}
\begin{table}[]
\centering
\resizebox{1.4\textwidth}{!}{%
\begin{tabular}{|c|c|cccccc|cccccc|cc|}
\hline
\multirow{3}{*}{Normal prediction model} & \multirow{3}{*}{Causal effect prediction model} & \multicolumn{6}{c|}{Original model}                                                                                                                                        & \multicolumn{6}{c|}{Two stage model}                                                                             & \multicolumn{2}{c|}{Improvement}                                       \\  
                                         &                                                 & \multicolumn{3}{c}{All OD}                                                                    & \multicolumn{3}{c|}{Influenced OD}                                        & \multicolumn{3}{c}{\begin{tabular}[c]{@{}l@{}}All OD\end{tabular}} & \multicolumn{3}{c|}{Influenced OD} & \multirow{2}{*}{All OD} & \multirow{2}{*}{Influenced OD} \\ 
                                         &                                                 & MAE                    & RMSE                   & \multicolumn{1}{l}{MAPE}                    & MAE                    & RMSE                   & MAPE                    & MAE              & RMSE             & \multicolumn{1}{l}{MAPE}             & MAE       & RMSE      & MAPE       &                                                    &           \\ \hline
\multirow{3}{*}{LR}                      & LR                                              & \multirow{3}{*}{2.177} & \multirow{3}{*}{3.344} & \multicolumn{1}{l|}{\multirow{3}{*}{14.6\%}} & \multirow{3}{*}{6.662} & \multirow{3}{*}{7.851} & \multirow{3}{*}{35.9\%} & 2.058            & 3.139            & \multicolumn{1}{l|}{13.6\%}           & 4.888     & 6.474     & 30.4\%     & 5.5\%                                                  & 26.6\%        \\
                                         & RF                                              &                        &                        & \multicolumn{1}{l|}{}                        &                        &                        &                         & 2.017            & 3.005            & \multicolumn{1}{l|}{13.1\%}           & 4.515     & 5.971     & 28.4\%     & 7.4\%                                                  & 32.2\%        \\ 
                                         & GBDT                                            &                        &                        & \multicolumn{1}{l|}{}                        &                        &                        &                         & 2.037            & 3.045            & \multicolumn{1}{l|}{13.3\%}           & 4.658     & 6.093     & 29.5\%     & 6.4\%                                                  & 30.1\%        \\ \hline 
\multirow{3}{*}{RF}                      & LR                                              & \multirow{3}{*}{1.902} & \multirow{3}{*}{2.771} & \multicolumn{1}{l|}{\multirow{3}{*}{12.5\%}} & \multirow{3}{*}{5.817} & \multirow{3}{*}{7.295} & \multirow{3}{*}{28.1\%} & 1.820            & 2.781            & \multicolumn{1}{l|}{12.8\%}           & 4.478     & 5.916     & 28.7\%     & 4.3\%                                                  & 23.0\%        \\
                                         & RF                                              &                        &                        & \multicolumn{1}{l|}{}                        &                        &                        &                         & 1.784            & 2.650            & \multicolumn{1}{l|}{12.4\%}           & 3.882     & 5.194     & 25.6\%     & 6.2\%                                                  & 33.3\%        \\
                                         & GBDT                                            &                        &                        & \multicolumn{1}{l|}{}                        &                        &                        &                         & 1.807            & 2.737            & \multicolumn{1}{l|}{12.9\%}           & 4.355     & 5.733     & 29.2\%     & 5.0\%                                                  & 25.1\%        \\ \hline
\multirow{3}{*}{GBDT}                    & LR                                              & \multirow{3}{*}{1.909} & \multirow{3}{*}{2.790} & \multicolumn{1}{l|}{\multirow{3}{*}{12.1\%}} & \multirow{3}{*}{5.869} & \multirow{3}{*}{7.378} & \multirow{3}{*}{27.6\%} & 1.824            & 2.897            & \multicolumn{1}{l|}{13.1\%}           & 4.427     & 5.315     & 27.9\%     & 4.5\%                                                  & 24.6\%        \\
                                         & RF                                              &                        &                        & \multicolumn{1}{l|}{}                        &                        &                        &                         & 1.779            & 2.646            & \multicolumn{1}{l|}{12.8\%}           & 3.995     & 5.302     & 26.5\%     & 6.8\%                                                  & 31.9\%        \\
                                         & GBDT                                            &                        &                        & \multicolumn{1}{l|}{}                        &                        &                        &                         & 1.808            & 2.743            & \multicolumn{1}{l|}{12.8\%}           & 4.368     & 5.760     & 27.7\%     & 5.3\%                                                  & 25.6\%        \\ \hline
\multirow{3}{*}{ST-ResNet}               & LR                                              & \multirow{3}{*}{1.872} & \multirow{3}{*}{2.703} & \multicolumn{1}{l|}{\multirow{3}{*}{10.5\%}} & \multirow{3}{*}{6.313} & \multirow{3}{*}{7.525} & \multirow{3}{*}{29.2\%} & 1.794            & 2.605            & \multicolumn{1}{l|}{12.3\%}           & 4.169     & 5.493     & 26.9\%     & 4.1\%                                                  & 34.0\%        \\
                                         & RF                                              &                        &                        & \multicolumn{1}{l|}{}                        &                        &                        &                         & 1.761            & 2.573            & \multicolumn{1}{l|}{12.2\%}           & 4.118     & 5.418     & 26.6\%     & 5.9\%                                                  & 34.8\%        \\
                                         & GBDT                                            &                        &                        & \multicolumn{1}{l|}{}                        &                        &                        &                         & 1.777            & 2.583            & \multicolumn{1}{l|}{12.3\%}           & 4.152     & 5.463     & 26.7\%     & 5.1\%                                                  & 34.2\%        \\ \hline
\multirow{3}{*}{U-Net}                   & LR                                              & \multirow{3}{*}{1.848} & \multirow{3}{*}{2.594} & \multicolumn{1}{l|}{\multirow{3}{*}{10.4\%}} & \multirow{3}{*}{6.274} & \multirow{3}{*}{7.689} & \multirow{3}{*}{31.2\%} & 1.787            & 2.702            & \multicolumn{1}{l|}{12.4\%}           & 4.287     & 5.673     & 27.7\%     & 3.3\%                                                  & 31.7\%        \\
                                         & RF                                              &                        &                        & \multicolumn{1}{l|}{}                        &                        &                        &                         & 1.730            & 2.605            & \multicolumn{1}{l|}{12.3\%}           & 4.169     & 5.493     & 26.9\%     & 6.4\%                                                  & 33.6\%        \\
                                         & GBDT                                            &                        &                        & \multicolumn{1}{l|}{}                        &                        &                        &                         & 1.741            & 2.663            & \multicolumn{1}{l|}{12.4\%}           & 4.236     & 5.568     & 27.3\%     & 5.8\%                                                  & 32.5\%        \\ \hline
\multirow{3}{*}{GEML}                    & LR                                              & \multirow{3}{*}{1.895} & \multirow{3}{*}{2.644} & \multicolumn{1}{l|}{\multirow{3}{*}{9.9\%}}  & \multirow{3}{*}{6.205} & \multirow{3}{*}{7.591} & \multirow{3}{*}{29.8\%} & 1.829            & 2.710            & \multicolumn{1}{l|}{12.5\%}           & 4.270     & 5.598     & 27.5\%     & 3.5\%                                                  & 31.2\%        \\
                                         & RF                                              &                        &                        & \multicolumn{1}{l|}{}                        &                        &                        &                         & 1.789            & 2.654            & \multicolumn{1}{l|}{12.4\%}           & 4.236     & 5.568     & 27.3\%     & 5.6\%                                                  & 31.7\%        \\
                                         & GBDT                                            &                        &                        & \multicolumn{1}{l|}{}                        &                        &                        &                         & 1.797            & 2.637            & \multicolumn{1}{l|}{12.4\%}           & 4.220     & 5.538     & 27.1\%     & 5.1\%                                                  & 32.0\%        \\ \hline
\end{tabular}
}
\caption{Experiment results}
\label{tab:result}
\end{table}

\end{landscape}
\restoregeometry
First, we summarize the performance of these baseline models without adjustments for causal effects. The errors on all passenger flows and significantly affected passenger flows (p-values $\leq$ 0.05) show that most models experience a substantial increase in error when predicting the flows of significantly affected ODs. In these cases, errors even surge by as much as three times. Traditional machine learning methods generally produce higher prediction errors across all flows compared to deep learning models. However, for significantly affected flows, traditional models exhibit lower prediction errors than deep learning models. This discrepancy likely occurs because deep learning models capture subtle and sensitive data patterns. However, these patterns often fail to generalize during incident conditions. Consequently, when deployed in such scenarios, the models exhibit rapid error increases.

Table \ref{tab:result} demonstrates that our two-stage method—first predicting with traditional models and then adjusting based on causal effects—significantly enhances accuracy, particularly for passenger flows that are significantly impacted. For overall passenger flows, we observe error reductions of approximately 4-7\%. In the case of incident-affected flows, error reductions reach 20-30\%. These findings validate the effectiveness of our approach.

Moreover, the results presented in Table \ref{tab:result} indicate that the Random Forest algorithm yields the most accurate predictions of causal effects, followed by Gradient Boosted Decision Trees, while Linear Regression offers the least improvement. This finding aligns with intuitive reasoning, as the impact of incidents on subway passenger flow is likely non-linear and may involve complex interactions. These mechanisms will be discussed in greater detail in the following section
\subsection{Analysis of causal effect prediction model}
In this section, we aim to answer a key question: how do these selected features influence the causal effects of incidents? Given that the Random Forest model demonstrated superior predictive accuracy, we conducted an in-depth analysis of the Random Forest model. First, we determine the importance of each feature in the Random Forest model, as shown in Table \ref{tab:importance}.
\renewcommand{\arraystretch}{1}
\begin{table}[!h]
\centering
\begin{tabular}{cc}
\toprule
Feature               & Importance \\ \midrule
$x_0$                    & 0.493      \\
time\_diff\_to\_start & 0.116      \\
time\_diff\_to\_end   & 0.076      \\
proportion            & 0.068      \\
max\_delay            & 0.056      \\
distance\_o           & 0.041      \\
distance\_d           & 0.034      \\
duration              & 0.015      \\
cancel\_num           & 0.013      \\
delay\_5\_num         & 0.012      \\
is\_in\_incident      & 0.010      \\
clear\_num            & 0.008      \\ \bottomrule
\end{tabular}
\caption{Feature importance}
\label{tab:importance}
\end{table}

Table \ref{tab:importance} shows that the counterfactual passenger flow ($x_0$) is the most significant element affecting the causal effect of incidents, which is consistent with common sense. This demonstrates that if an OD has a higher baseline passenger flow, an incident will have a greater influence on it. In contrast, for OD with smaller baseline flow, the incident exhibits a modest effect on it. Other key factors include the time from the start and end of the incident to the time of OD, which also corresponds to common expectations. Short intervals leave insufficient time for impacts to manifest, while long intervals allow effects to dissipate. Therefore, the time interval is crucial to the effect of the incident. Additionally, the delay duration caused by the incident is also highly important, which makes sense, as longer delays generally reflect more severe incidents, and therefore, a greater impact on passenger flow.

However, features such as the number of canceled trains and the number of five-minute delay trains appear less significant. This might be because the severity of incidents does not directly correspond to these metrics. For instance, the count of five-minute delay trains might increase as the incident becomes more serious, but as the severity intensifies further, some delayed trains might be canceled, causing the number to decrease. The lack of a one-to-one relationship suggests that this metric may not clearly reflect the severity of incidents. These findings indicate that future models may benefit from including indicators that reflect the incident severity more accurately.
\begin{figure}[h]
    \centering
    \includegraphics[width=\linewidth]{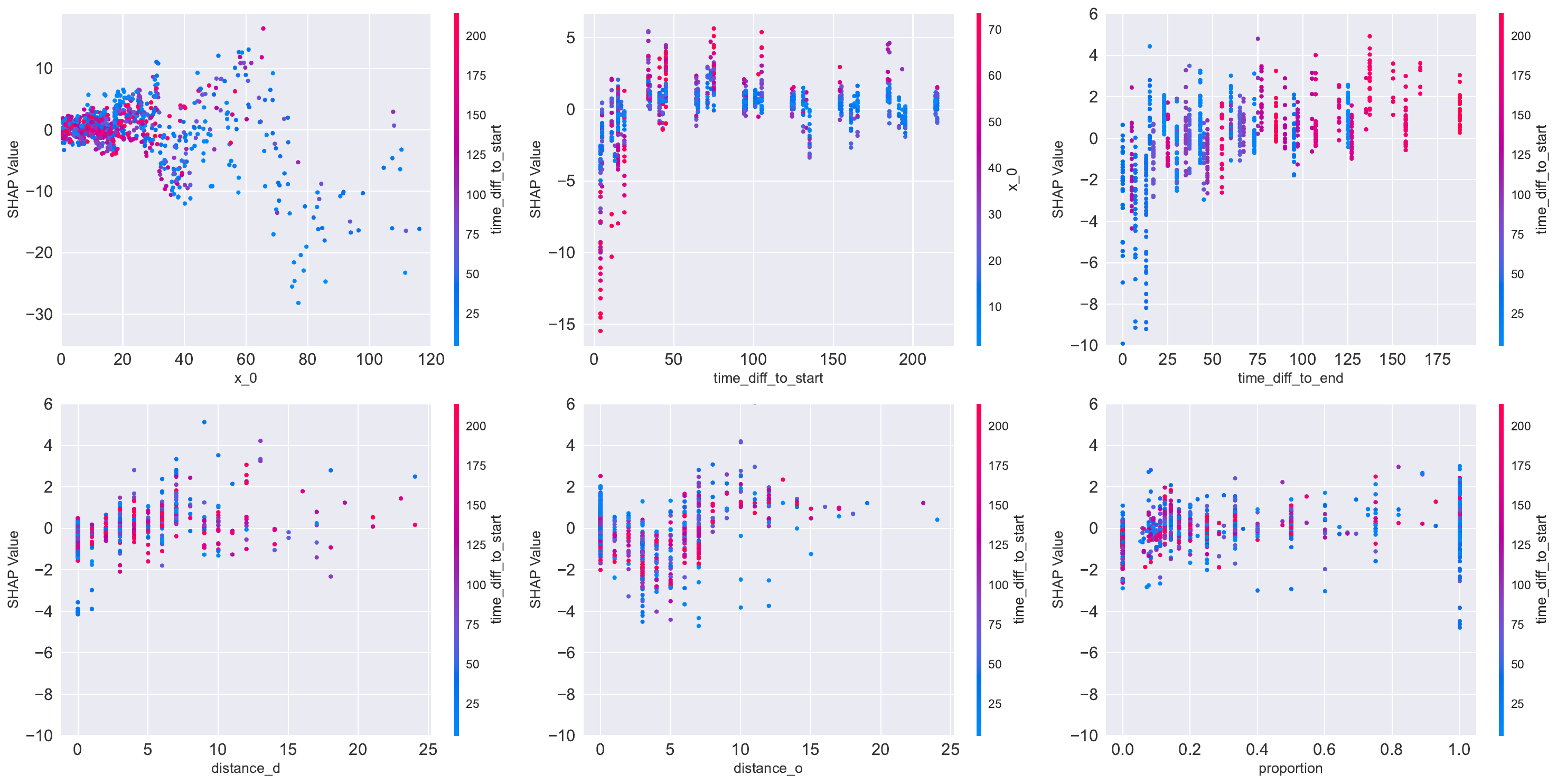}
    \caption{Partial dependence plots for some features}
    \label{fig:pdp}
\end{figure}

Besides, we conducted ablation experiments to validate the importance of each feature group on causal effect prediction. The results are represented in Tab \ref{tab:abl}. The ablation study reveals that removing any feature group degrades model accuracy; the most significant performance degradation occurs when excluding counterfactual passenger flow features. This aligns with our feature importance analysis, confirming that counterfactual passenger flow ($x_0$) is the most critical predictor of incident impacts.

\begin{table}[h]
\begin{tabular}{@{}cccc@{}}
\toprule
                                        & R2    & MAE   & RMSE  \\ \midrule
Ori model                               & 0.550  & 4.909 & 7.605 \\
w\textbackslash{}o incident severity    & 0.494 & 5.691 & 8.206 \\
w\textbackslash{}o temporal correlation & 0.484 & 5.552 & 8.288 \\
w\textbackslash{}o spatial correlation  & 0.482 & 5.486 & 8.306 \\
w\textbackslash{}o $x_0$                     & 0.425 & 5.788 & 8.834 \\ \bottomrule
\end{tabular}
\caption{Ablation experiment results}
    \label{tab:abl}
\end{table}
Moreover, we analyzed model outputs by plotting partial dependence plots (PDPs) using the SHAP method \cite{scott2017unified}. Each subplot in Figure \ref{fig:pdp} represents a PDP and shows the interaction between two variables, where the x-axis indicates one variable and the color represents another. For example, in the first subplot, larger counterfactual baseline flows ($x_0$) generally correlate with stronger causal effects. However, the signs of the causal effect vary by color, which reflects the time difference between the OD and the start of the incident. Smaller time differences (bluer points) correlate with negative causal effects, indicating suppressed flow, while larger time differences (redder points) suggest increased flow, as some users delay travel and lead to a recovery effect.

Similar patterns appear in the second subplot in the first row, where causal effects are predominantly negative when time differences are below 30 min, turning positive at around 50-100 minutes, possibly indicating flow recovery. After 100 minutes, causal effects diminish around zero, and negative effects tend to be stronger than positive ones. This implies that a subway incident could lead to an initial sharp drop in passenger flow, followed by gradual recovery. Even after the incident ends, the flow may remain slightly above the normal levels.

The first and second subfigures in the second row illustrate how the distance from the incident location to the origin and destination stations affects the causal effects. The effect is generally stronger when the distance is within approximately 10, but it diminishes significantly beyond this range. When the distance is less than 10, the absolute values of the causal effect are generally small, suggesting a threshold effect of distance. Additionally, the distance to the origin station has a more pronounced influence on the causal effect than the distance to the destination station.
\subsection{Validation of sample selection}
In the method section, we proposed that it is important to select samples that are significantly impacted to calibrate causal effect prediction model. In deployment, choosing passenger flows that are significantly affected by incidents is also beneficial. To validate this statement, and to analyze how different threshold values ($p1$ and $p2$ in Figure \ref{fig:wolkflow_detail}) affect overall prediction error, we conducted further analysis.

\begin{figure}[h]
    \centering
    \includegraphics[width=\linewidth]{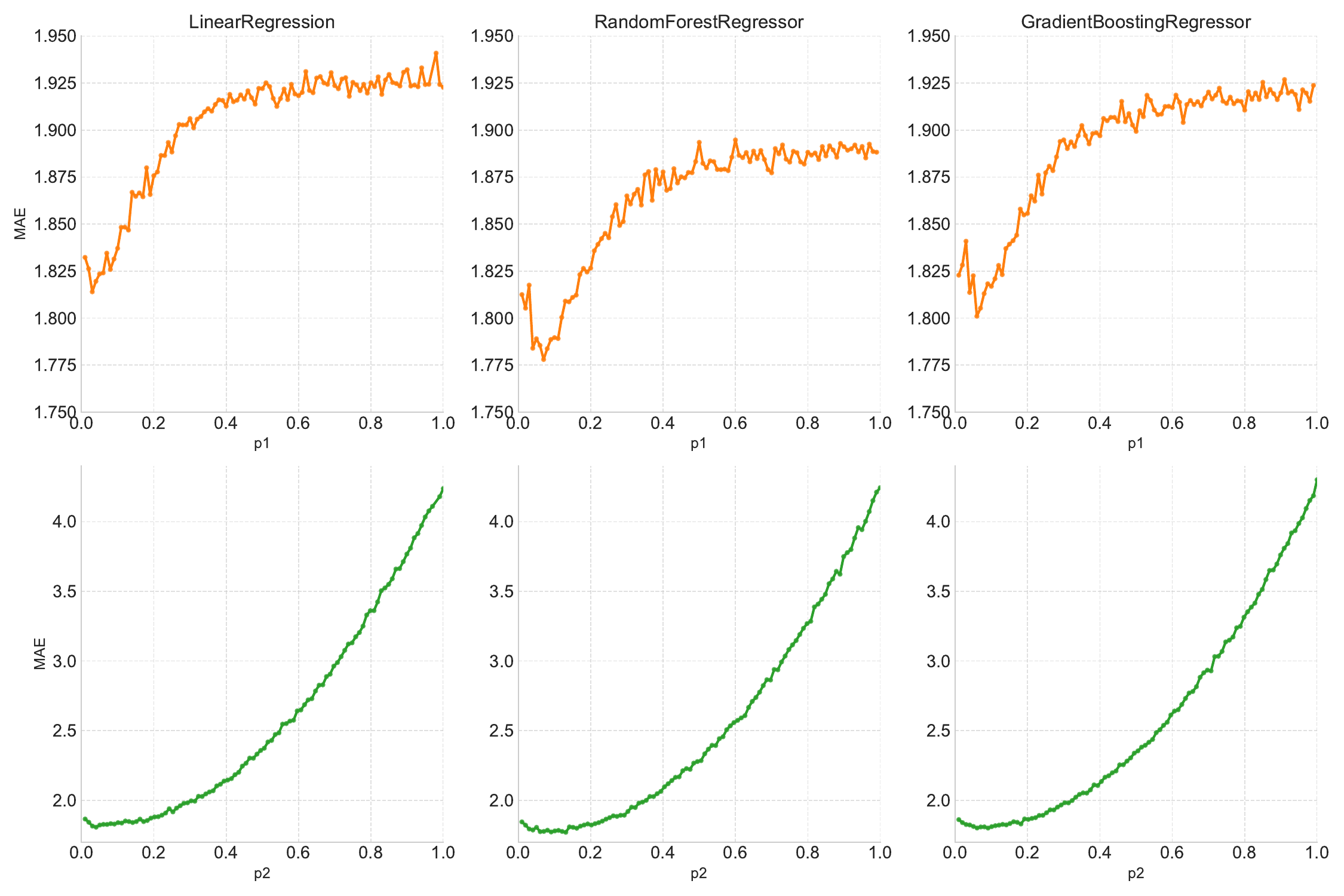}
    \caption{The relation between $p1,p2$ and prediction errors}
    \label{fig:mae-p}
\end{figure}

Specifically, we used Random Forest as the baseline model under normal situation. First, we fixed $p1$ at 0.05 and adjusted $p2$ to observe how changes in $p2$ impact overall prediction error. We examined three causal effect prediction models: Linear Regression, Random Forest, and Gradient Boosted Decision Trees, producing the three subplots in the lower half of Figure \ref{fig:mae-p}.

Next, we fixed $p2$ at 0.05 and adjusted $p1$, varying it between 0 and 1. For each setting, we repeated the model calibration and deployment processes, measuring the final overall passenger flow prediction accuracy. We again used Linear Regression, Random Forest, and Gradient Boosted Decision Trees as causal effect prediction model, respectively. The results are shown in the top three subplots in Figure \ref{fig:mae-p}.

From the top three subplots in Figure \ref{fig:mae-p}, we observe that prediction error is high when $p1$ is set either very low (close to zero) or very high (close to one). The lowest prediction errors occur when $p1$ is in the middle range, around 0.03 to 0.08, which aligns with previous theoretical expectations. Comparing the three prediction models, we see that Linear Regression achieves optimal prediction accuracy with a smaller $p1$ value, around 0.03, while Random Forest and Gradient Boosted Decision Trees require a slightly larger $p1$.

This finding aligns with prior analysis: the final prediction error is determined by both the quantity and quality of selected samples. For the linear model, which has fewer parameters compared to tree-based models, fewer samples are needed to achieve adequate calibration. This allows for a smaller threshold ($p1$), ensuring higher-quality samples are selected.

Observing the lower half of Figure \ref{fig:mae-p}, it can be found that when $p2$ is either too low, such as zero, or too high, close to one, the prediction error of the model is relatively large. Only when $p2$ falls within a reasonable range, approximately 0.1, does the model achieve optimal prediction accuracy. Additionally, the relationship between $p2$ and the final prediction error resembles a quadratic function, which aligns well with the result of Theorem \ref{thro:adjust}. This similarity supports the validity of our theoretical findings.

A cross-model comparison also reveals that the optimal $p2$ value for Linear Regression is smaller compared to the other two models. This observation is consistent with previous analysis, which suggests that models with higher goodness of fit in causal effect prediction need a larger $p2$ threshold. Complex tree-based models, such as Random Forest and Gradient Boosted Decision Trees, exhibit better fit than Linear Regression, allowing them to adopt higher $p2$ values for sample selection.
\section{Conclusion}

In this paper, we propose a two-stage method to predict subway passenger flows under incident conditions. In the training phase, we first use synthetic control methods to evaluate the causal effects of incidents on passenger flows, followed by calibrating a model using incident and passenger flow features to predict these causal effects. And a prediction model for normal situation is trained. In the deployment phase, the normal model predicts passenger flows without incidents. Then, the causal effect prediction model is used to estimate the effects of the incident, and the two results are combined to obtain the final prediction. We also use placebo test method to obtain the significant level of causal effect and the importance of it is validated both theoretically and experimentally. The experimental results based on real-world data demonstrate the effectiveness of the proposed method.

The two-stage style provides advantages in interpretability and flexibility. In terms of interpretability, the model allows for an analysis of incident’s effects, identifying how the impact of incidents spreads and dissipates, and how passenger flows are suppressed and then gradually recover. In terms of flexibility, different models can be used for normal flow prediction and causal effect prediction, with the potential for improved accuracy using more advanced models in the future.

There are several limitations in our current work which could be improved. First, the current 30-minute prediction interval introduces latency in emergency response. Future work should develop sparse-data tolerant models (e.g., zero-inflated regression) and take use of larger dataset for finer-grained prediction. Besides, the reliance on post-incident features (e.g., maximum delay time) impedes real-time deployment. Subsequent studies need to use immediately available incident descriptors (e.g., incident severity level, incident type) for causal effect estimation. Moreover, our model treats ODs as independent units, neglecting network-wide behavioral interdependencies. Collecting larger-scale incident datasets would enable spatiotemporal graph-based approaches to capture passenger rerouting dynamics.

Another major limitation lies in the dataset: it only contains six eligible incidents—primarily caused by signal failures, mechanical and electrical equipment malfunctions. However, during extreme events like fires or terrorist attacks, passenger behavioral responses would diverge significantly from those observed in our dataset. This restricts our model's generalizability. Future studies with datasets encompassing diverse incident types may address this by calibrating separate models for distinct emergency categories.

Finally, forecasting passenger flow under special conditions remains challenging due to limited data on such scenarios. When collecting sufficient data proves difficult, large language models (LLMs) could offer an alternative, as recent research has shown that LLMs, with their extensive world knowledge and generalization ability, can make reasonable predictions for unusual scenarios \cite{Huang2024EnhancingTP,liang2024exploring}. Integrating LLMs for subway passenger flow prediction under incident conditions may be a promising future direction.

In summary, our work can help subway operators predict changes in passenger flow after incidents, providing input for designing alternative transportation or implementing emergency measures. It can also deepen researchers’ understanding of the mechanisms behind the effects of incidents on passenger flows.
\section*{Author contributions}
Conceptualisation: Chao Yang, data curation: Xiannan Huang, methodology: Chao Yang+Xiannan Huang+Shuhan Qiu, software: Xiannan Huang, formal analysis: Xiannan Huang, investigation: Xiannan Huang+Quan Yuan, writing-original draft: Xiannan Huang, writing-review and editing: Shuhan Qiu+Quan Yuan+Chao Yang, visualisation: Xiannan Huang, supervision: Chao Yang. All authors reviewed the results and approved the final version of the manuscript.
\section*{Statements and Declarations}
\subsection*{Conflict of interest}
The authors declare no competing interests.
\bibliography{ref}
\appendix
\section{Proof of Theorem \ref{theo:3.1}}
\textbf{Theorem3.1}

If there are n samples obeying the linear model with noised samples in definition \ref{noisy-sample} and we fit a model with these samples, resulting in the fitted parameter  $\hat{\beta}$. Then, the square distance between true parameter and the fitted parameter is bounded by:
\begin{equation}
E(\beta-\hat{\beta})^2=(1-p)\ |\beta|^2+c\frac{(p\sigma_2^2+\sigma_1^2)}{n}
\end{equation}

\textbf{Proof:}

Suppose $X\in R^{n\times d},Y\in R^n,\beta\in R^d$, where $n$ is sample size and $d$ is the dimension of feature. We first show that the $l_2$ loss can be decomposed to the summation of bias and variance of $\hat{\beta}$, then we tackle both bias and variance.

1) bias and variance decomposition:
\begin{align}
E\left(\beta-\hat{\beta}\right)^2 &= 
E\left(\beta-E\hat{\beta}+E\hat{\beta}-\hat{\beta}\right)^2 \\
&= E\left(\beta-E\hat{\beta}\right)^2 + E\left(E\hat{\beta}-\hat{\beta}\right)^2 
+ 2E\left[\left(\beta-E\hat{\beta}\right)\left(E\hat{\beta}-\hat{\beta}\right)\right]
\end{align}
Because $E[\left(\beta-E\hat{\beta}\right)\left(E\hat{\beta}-\hat{\beta}\right)=\left(\beta-E\hat{\beta}\right)E\left(E\hat{\beta}-\hat{\beta}\right)=\left(\beta-E\hat{\beta}\right)\times\left(E\hat{\beta}-E\hat{\beta}\right)=0$
Then:
\begin{equation}
    E\left(\beta-\hat{\beta}\right)^2=E\left(\beta-E\hat{\beta}\right)^2+E\left(E\hat{\beta}-\hat{\beta}\right)^2=bias\left(\hat{\beta}\right)^2+Tr(Var\left(\hat{\beta}\right))
\end{equation}

2) bias term:
\begin{equation}E\left(\hat{\beta}\right)=E\left[\left(XX^T\right)^{-1}X^TY\right]=\left(XX^T\right)^{-1}X^TE\left(Y\middle| X\right)
\end{equation}
Then:
\begin{equation}E\left(Y\middle| X\right)=(1-p)\times E\epsilon_1+pE\left(\epsilon_1+\left(\beta^TX+\epsilon_2\right)\right)=pX\beta
\end{equation}
Therefore:
\begin{equation}E\left(\hat{\beta}\right)=E\left(1-p\right)\left(XX^T\right)^{-1}X^TX\beta=p\beta
\end{equation}
So:
\begin{equation}
bias\left(\hat{\beta}\right)^2=\left((1-p)\beta\right)^2
\end{equation}

3) variance term:
\begin{equation}
Var\left(\hat{\beta}\right)=Var\left(\left(XX^T\right)^{-1}X^TY\right)=\left(XX^T\right)^{-1}X^TVar\left(Y\middle| X\right)X\left(XX^T\right)^{-1}
\end{equation}
Then:
\begin{equation}
Var\left(Y\middle| X\right)=\left((1-p)\sigma_1^2+p\left(\sigma_1^2+\sigma_2^2\right)\right)I_n=\left(\sigma_1^2+p\sigma_2^2\right)I_n
\end{equation}
Where $I_n$ is $n\times n$ identical matrix. Substituting it into $Var\left(\hat{\beta}\right)$
\begin{equation}
Var\left(\hat{\beta}\right)=\left(XX^T\right)^{-1}X^T\left(\sigma_1^2+p\sigma_2^2\right)I_nX\left(XX^T\right)^{-1}
\end{equation}
Then:
\begin{equation}
Tr\left(Var(\hat{\beta})\right)=\left(\sigma_1^2+p\sigma_2^2\right)Tr(\left(XX^T\right)^{-1})
\end{equation}
Suppose each entry in $X$ is i.i.d sample from $N\left(0,\sigma_x^2\right)$, then $XX^T$ is Wishart distribution \cite{bilodeau1999theory} and:
\begin{equation}
E{Tr\left(XX^T\right)}^{-1}=\frac{d}{(n-d-1)\sigma_x^2}
\end{equation}
Then:
\begin{equation}
Tr\left(Var\left(\hat{\beta}\right)\right)=\left(\sigma_1^2+p\sigma_2^2\right)\frac{d}{(n-d-1)\sigma_x^2}
\end{equation}
For large $n$:
\begin{equation}
Tr\left(Var\left(\hat{\beta}\right)\right)\approx\left(\sigma_1^2+p\sigma_2^2\right)\frac{d}{n\sigma_x^2}
\end{equation}

4) integrate two terms:
\begin{equation}
E\left(\beta-\hat{\beta}\right)^2={(1-p)}^2\beta^2+\left(\sigma_1^2+p\sigma_2^2\right)\frac{d}{n\sigma_x^2}={(1-p)}^2\beta^2+c\frac{\sigma_1^2+p\sigma_2^2}{n}
\end{equation}
where:
$c=\frac{d}{\sigma_x^2}$
\section{Proof of Theorem \ref{thro:adjust}}

We first elaborate this theorem in detail again.

\textbf{Theorem3.1}

If the influence factors about the causal effects is noted as $x$, the causal effect is donated as $y$, assume $y$ and $x$ obeys the following rule:
$$y=\epsilon_1\ with\ probability\ 1-p$$
$$y=\epsilon_1+(f(x)+\epsilon_2)\ with\ probability\ p$$
Where $\epsilon_1$ following normal distribution $  N(0,\sigma_1^2)$ and $\epsilon_2$ following normal distribution  $N(0,\sigma_2^2)$, $\beta$ is the parameter and p is a constant between 0 and 1. 

If we have observed a lot of $\left(x_i,p_i\right)$ and a prediction model $\hat{y}=\hat{f}(x)$
And we have a threshold $P$, for each OD, if $p_i$, is greater than $P$, we will use causal effect prediction model to adjust it, otherwise we will just regard the prediction of normal model as the final prediction value. Then we can get the risk of this strategy.
\begin{equation}
E\left(y-\hat{y}\right)^2=\frac{1}{2}P^2(Ef\left(x\right)^2+E\hat{f}\left(x\right)^2-E\left(\hat{f}\left(x\right)-f\left(x\right)\right)^2)-PE\hat{f}\left(x\right)^2+c
\end{equation}

\textbf{proof:}

There are 4 cases: the OD is truly affected, but we not adjust; the OD is truly affected and we adjust it; the OD is not truly affected, and we not adjust; the OD is not truly affected but we adjust it; We need add the risk of these four cases:
\begin{align}
E\left(y - \hat{y}\right)^2 = & \; E \int P\left(p_i > P \ \text{and} \ y_i = \epsilon_1 + \left(f(x_i) + \epsilon_2\right)\right) \\
&\times \left(\hat{f}(x_i) - \left(\epsilon_1 + \left(f(x_i) + \epsilon_2\right)\right)\right)^2 \, dp_i \\
& + \int P\left(p_i > P \ \text{and} \ y_i = \epsilon_1\right) \left(\hat{f}(x_i) - \epsilon_1\right)^2 \, dp_i \\
& + \int P\left(p_i < P \ \text{and} \ y_i = \epsilon_1\right) \epsilon_1^2 \, dp_i \\
& + \int P\left(p_i < P \ \text{and} \ y_i = \epsilon_1 + \left(f(x_i) + \epsilon_2\right)\right) \left(\epsilon_1 + \left(f(x_i) + \epsilon_2\right)\right)^2 \, dp_i
\end{align}

The first term in the risk when the OD is affected and we adjust it, the second is the risk when the OD is not affected but we adjust it.
Because $P\left(y_i=\epsilon_1+\left(\beta^Tx+\epsilon_2\right)\right)=p_i$ and $P\left(y_i=\epsilon_1\right)={1-p}_i$, and we further need to assume $p_i$ obey uniform distribution in [0,1]. then the above expression equals to:
\begin{align}
E\left(y - \hat{y}\right)^2 = & \; E \Big[ \int_{P}^{1} t \left( \hat{f}(x) - \left( \epsilon_1 + \left( f(x) + \epsilon_2 \right) \right) \right)^2 dt  + \int_{P}^{1} (1 - t) \left( \hat{f}(x) - \epsilon_1 \right)^2 dt \\
& + \int_{0}^{P} t \left( \epsilon_1 + \left( f(x) + \epsilon_2 \right) \right)^2 dt + \int_{0}^{P} (1 - t) \epsilon_1^2 dt \Big]
\end{align}

Because $\epsilon_1$ follows normal distribution $ N(0,\sigma_1^2)$ and $\epsilon_2$ follows normal distribution $N(0,\sigma_2^2)$ and we further assume they are independent. Therefore, we have:
$$
Ef(x)\epsilon_1=0,\ Ef(x)\epsilon_2=0,\ E\epsilon_1\epsilon_2=0
$$

Then by some calculation, we can get: 
\begin{align}
E\left(y - \hat{y}\right)^2 = & \; \frac{1}{2}P^2 \left(Ef\left(x\right)^2 + E\hat{f}\left(x\right)^2 
- E\left(\hat{f}\left(x\right) - f\left(x\right)\right)^2\right) \\
& - P E\hat{f}\left(x\right)^2 
+ E\left(\hat{f}(x)^2+\frac{f\left(x\right)^2}{2} - f\left(x\right)\hat{f}\left(x\right)\right) 
+ \sigma_1^2 + \frac{\sigma_2^2}{2} \\
= & \; \frac{1}{2}P^2 \left(Ef\left(x\right)^2 + E\hat{f}\left(x\right)^2 
- E\left(\hat{f}\left(x\right) - f\left(x\right)\right)^2\right)  - P E\hat{f}\left(x\right)^2 + c
\end{align}

where:
$$c=E\left(\hat{f}(x)^2+\frac{f\left(x\right)^2}{2}-f\left(x\right)\hat{f}\left(x\right)\right)+\sigma_1^2+\frac{\sigma_2^2}{2}$$ 
\end{document}